\documentclass[aip,
 amsmath,amssymb,
 preprint,
]{revtex4-2}

\usepackage{graphicx}% Include figure files
\usepackage{dcolumn}% Align table columns on decimal point
\usepackage{bm}% bold math
\usepackage[utf8]{inputenc}
\usepackage[T1]{fontenc}
\usepackage{mathptmx}
\usepackage{etoolbox}
\usepackage{comment}
\usepackage{amsmath, amssymb}
\usepackage{float}

\DeclareMathOperator*{\argmax}{arg\,max}
\DeclareMathOperator*{\argmin}{arg\,min}

\makeatletter
\def\@email#1#2{%
 \endgroup
 \patchcmd{\titleblock@produce}
  {\frontmatter@RRAPformat}
  {\frontmatter@RRAPformat{\produce@RRAP{*#1\href{mailto:#2}{#2}}}\frontmatter@RRAPformat}
  {}{}
}%
\makeatother
\begin{document}

\preprint{AIP/123-QED}

\title[UQ of NN Turbulence Closures]{Quantifying Out-of-Training Uncertainty of Neural-Network based Turbulence Closures}
% Force line breaks with \\
\author{C. Grogan}
 \affiliation{Mechanical and Aerospace Engineering, Utah State University}
\email{A02313514@usu.edu.}
 
\author{S.L.N. Dhulipala}%
\affiliation{Idaho National Laboratory}%

\author{M. Tano}%
\affiliation{Idaho National Laboratory}%

\author{I. Gutowska}%
\affiliation{School of Nuclear Science and Engineering, Oregon State University
}%

\author{S. Dutta}
% \homepage{http://www.Second.institution.edu/~Charlie.Author.}
\affiliation{Mechanical and Aerospace Engineering, Utah State University}
\email{som.dutta@usu.edu.}

\date{\today}% It is always \today, today,
             %  but any date may be explicitly specified

\begin{abstract}
Neural-Network (NN) based turbulence closures have been developed for being used as pre-trained surrogates for traditional turbulence closures, with the aim to increase computational efficiency and prediction accuracy of CFD simulations. 
One of the bottlenecks to the widespread adaptation of these ML-based closures is the relative lack of uncertainty quantification (UQ) for these models. Especially, quantifying uncertainties associated with out-of-training inputs, that is when the ML-based turbulence closures are queried on inputs outside their training data regime.
In the current paper, a published algebraic turbulence closure \cite{taghizadeh2021turbulence} has been utilized to compare the quality of epistemic UQ between three NN-based methods and Gaussian Process (GP). 
%which is well known in the uncertainty quantification literature for giving principled uncertainty estimates.
%
The three NN-based methods explored are Deep Ensembles (DE), Monte-Carlo Dropout (MCD), and Stochastic Variational Inference (SVI).
In the in-training results, we find the exact GP performs the best in accuracy with a Root Mean Squared Error (RMSE) of $2.14 \cdot 10^{-5}$ followed by the DE with an RMSE of $4.59\cdot 10^{-4}$.
For UQ accuracy on the in-training results, the DE exhibits the lowest miscalibration error and a negative log-likelihood close to the more computationally expensive Exact GP.
Next, the paper discusses the performance of the four methods for quantifying out-of-training uncertainties.
For performance, the Exact GP yet again is the best in performance, but has similar performance to the DE in the out-of-training regions.
In UQ accuracy for the out-of-training case, SVI and DE hold the best miscalibration error for one of the cases.
However, the DE performs the best in Negative Log-Likelihood for both out-of-training cases.
We observe that for the current problem, in terms of accuracy $GP>DE>SVI>MCD$. The DE results are relatively robust and provide intuitive UQ estimates, despite performing \textit{naïve ensembling}. In terms of computational cost, the GP is significantly higher than the NN-based methods with a $O(n^3)$ computational complexity for each training step. 
Thus, DE might be the most viable approach for UQ in large-scale NNs, that have high-dimensional inputs and complex input-output relationships, used for modeling turbulence closures.
\end{abstract}

\maketitle

%%%%%%%%%%%%%%%%%%%%%%%%%%%%%%%%%%%
\section{Motivation}
%%%%%%%%%%%%%%%%%%%%%%%%%%%%%%%%%%%
Machine learning (ML) models are receiving significant interest for accelerating fluid dynamics and transport related modeling and simulations \cite{li2022,el2021}. The ML and Deep-Learning based models have applied to different fields, including hemodynamics\cite{zou2025residual}, oil flow in reservoirs\cite{niu2025physics}, natural convection\cite{jangid2025application}, sediment incipient motion\cite{zhang2025data}, particle drag coefficient\cite{tian2025physics}, among others.  
Among the many applications of machine learning (ML) in fluid dynamics, one of the most impactful and rapidly advancing areas is its integration within computational fluid dynamics (CFD)\cite{vinuesa2022enhancing}.
CFD is a powerful numerical approach for simulating fluid flows, but its accuracy and computational cost often hinge on how turbulence is modeled. Turbulence is inherently complex and difficult to resolve directly, so CFD traditionally relies on approximate models derived from solving one, two, or many partial differential equations (PDEs) for surrogate turbulence fields, e.g., the turbulent kinetic energy and its dissipation rate in the standard $k-\epsilon$ model~\cite{launder1974}. Recently, ML has been leveraged to improve this aspect of CFD through the development of turbulence model emulators that can harvest larger datasets to provide tailored turbulent closures. In this approach, a pre-trained ML model is coupled with the CFD solver to provide faster—and in some cases more accurate—predictions of turbulent flow \cite{girimaji2024turbulence}, compared to conventional PDE-based turbulence closures \cite{fang2024cfd,rasmussen2021cfd,lloyd2023multi,tano2019development}.
Within the spectrum of ML approaches for turbulent closures, Neural-Network (NN)-based models have demonstrated promising performance for learning turbulence closure~\cite{mcconkey2022}. NN are a class of models that mathematically replicate the connection between neurons within a brain, and is the basis for more of the most sophisticated advancements in AI.
NN are universal approximators~\cite{scarselli1998} that can theoretically learn any function, including the turbulence closure models from a subset of high-fidelity experimental or numerical data.
Hence, NN-based turbulence closures can replace or increase the fidelity of turbulence closures used in large eddy simulations (LES) \cite{yuan2020deconvolutional} or Reynolds Average Navier Stokes (RANS) \cite{chang2020reynolds,xie2021artificial} based CFD simulations.

There are different approaches by which NN-closures can be implemented in lower-fidelity models.
One way of using NN-based models to increase the fidelity of RANS and LES is through the approach by Iskhakov et al.\cite{iskhakovmachine} for coarse-grid RANS and LES. Coarse-grid simulations for RANS and LES are special cases of each approach where the flow is filtered to exclude small-scale components of the flow.
%
%
%This yields equations for a fast-flow simulation at the cost of deriving complicated relations that account for small-scale contributions or using a closure.
%
The researchers in this article use traditional turbulence closures for the simulations but using a NN model to predict the eddy viscosity. Hence, the NN model acts as an error correction for the predicted flow velocity fields in the coarse-grid simulations against higher-fidelity data.
Iskhakov et al. show that the NN corrected versions of the coarse-grid flow simulations had a lower MSE than the coarse-grid baseline for the flow scenario explored in the study. 
This study showed the potential of using NN-based models to improve the accuracy of coarse-grid RANS and LES, while keeping the computational cost lower than the high-fidelity CFD simulations at computational grid resolution that will be required to achieve requisite accuracy.

A different approach for Neural Network-based turbulence closure is to directly model the Reynold's stress tensor using a NN-based model.
An example of this approach is presented by Ling \cite{ling2016reynolds}. In this approach, coined a Tensor Basis Neural Network (TBNN), a Neural Network is employed to determine the anisotropy tensor $b_{ij}$ using a NN with embedded Galilean invariance.
Galilean invariance is accomplished by expressing the anisotropy tensor in terms of a basis of tensors dependent only on the strain (\textbf{S}) and rotation rate (\textbf{R}) tensors, i.e.,  \textbf{T}(\textbf{S}, \textbf{R}) as follows:
\begin{align}
    \textbf{b} = \sum_{n=1}^{10} g^n(\lambda_1, ..., \lambda_5) T^n
    \label{eq:tensor_basis}
\end{align}
where $\textbf{b}$ is the anisotropy tensor, $\lambda_i$ are invariants, and $T^n$ are the basis tensors. Note that the Galilean is variance is achieved as none of the closure terms explicitly depends on space.
The complete mathematical formulation of the invariants $\lambda_i$ and basis tensors $T^n$ can be found in Ling\cite{ling2016reynolds}.
Ling \cite{ling2016reynolds} then uses a NN to predict the scalar projection coefficients $g^n$ of the basis tensors in Eq.~\ref{eq:tensor_basis} to yield more accurate RANS simulations when compared to DNS results.
In their results, Ling \cite{ling2016reynolds} shows the TBNN significantly out-performs all other methods in terms of accuracy of the true anisotropy tensor $\textbf{b}$ compared to the DNS baseline used for training the NN models.
This shows Neural Networks have the ability to play a significant role in RANS turbulence closures by directly modeling the Reynold's stress tensor.

Approaches for ML-augmented RANS and LES models of turbulence have been reviewed by Duraisamy~\cite{duraisamy2021perspectives}, providing a comprehensive perspective on the subject. Furthermore,Girimaji~\cite{girimaji2024turbulence}, in their recent perspective on turbulence closure modeling with ML, discusses the issues with the currently utilized approaches. One of such issues is that most ML modeling approaches disregard important physical principles and
mathematical constraints, leading to substantial decline in performance for unseen flows, i.e., flow conditions not present in the training set of the turbulent closure. Thus, quantification of this prediction deterioration is important for the wider community to trust and adopt these models.

Adding to these arguments, for engineering practice and regulatory compliance, one of the main impediments to the wider adoption of these ML models is the lack of systematic quantification of uncertainties in these models \cite{wu2025uncertainty}, especially when they operate in an extrapolated state. For ML-based models, an extrapolated state is when the model is queried outside the range of its training data, increasing the possibility of large prediction errors. This uncertainty in the prediction of ML models in an extrapolated state is also known as \textit{out-of-training} uncertainty \cite{gawlikowski2023survey}. Out-of-training uncertainty of ML models is intertwined with model/epistemic uncertainty associated with the NN models. This point will be further illustrated in the current paper. 

In the realm of fluid dynamics, relatively few works have incorporated uncertainty quantification (UQ) techniques into data-driven and deep learning models \cite{foglia5295305empirical}.
For example, using a TBNN network structure, recently, Man~\cite{man2025uncertainty} estimates the distribution over the anisotropy components by approximating them as Gaussian distributions. This allows one to estimate the uncertainty in the predictions of the anisotropy components of the Reynolds stress tensor, but may not reflect the true uncertainty of the NN model.
Geneva~\cite{Geneva_2020} uses Stochastic Variational Inference, a Bayesian inference technique, to quantify the uncertainty in model predictions due to the model parameters, a form of epistemic uncertainty. Also, as a precursor to the current study, the authors of this work quantified the model/epistemic uncertainty of NN-based turbulence closures, trained using data from an algebraic Reynolds stress model \cite{grogan2024quantifying}. In this study, different Bayesian UQ methods were explored.

The objective of the current paper is to expand the previous Bayesian UQ methods by comparing and contrasting three different NN-based methods and one traditional method for quantifying model uncertainty of a turbulence closure surrogate, where the model is developed using data from an algebraic turbulence closure \cite{taghizadeh2021turbulence}. The three NN-based methods for quantifying uncertainty are Deep Ensembles, Monte Carlo Dropout, and Stochastic Variational Inference, while Gaussian Processes is utilized as the traditional method.
The current paper is one of the first papers to illustrate the ``\textit{why, how, and what}" of uncertainty quantification of NN-based turbulence closures.

The rest of this article is organized as follows. In Section~\ref{sec::nn-closure}, the NN-based turbulent closure utilized in this work is described. Then, in Section~\ref{sec::nn-uq}, the framework for UQ in turbulence closure models is explained. The materialization of this framework via Deep Ensembles, Monte Carlo Dropout, Stochastic Variational Inference, and Gaussian Processes is explained in Section~\ref{sec::nn-bi}. Next, Section~\ref{sec::nn-results} presents the performance of these methods for in- and out-training UQ. Afterwards, these results are analyzed in further detail in Section~\ref{sec::nn-disc}. Finally, the main conclusions and future perspectives of this work are provided in Section~\ref{sec::conc}.

%%%%%%%%%%%%%%%%%%%%%%%%%%%%%%%%%%%%%%%%%%%%%%%%%
\section{NN-based Turbulence closure}
\label{sec::nn-closure}
%%%%%%%%%%%%%%%%%%%%%%%%%%%%%%%%%%%%%%%%%%%%%%%%%
Machine-learning (ML) models have shown the potential to significantly improve the accuracy of turbulence closures used for LES and RANS-based CFD simulations \cite{duraisamy2019turbulence,duraisamy2021perspectives}.
The ML models work on the rationale that the shortcomings of the physics-based closures can be mitigated by appropriate data–based training of the models. 
Among the different methods within ML, NN-based approaches have gained substantial traction, especially for RANS turbulence closure. Deep NN (DNN) with embedded invariance \cite{ling2016reynolds}, Bayesian DNN \cite{Geneva_2020} and other variations of DNN have shown promising results for different flow applications \cite{iskhakovmachine,volpiani2021machine}. 
Though, generalization of NN–based RANS modeling is difficult due to complex closure relations stemming from flow-dependent non–linearity and bifurcations, and difficulty in acquiring high–fidelity data covering all the regimes of relevance. 
To get over these impediments, 
Taghizadeh et al. \cite{taghizadeh2021turbulence} proposed conducting a preliminary analysis of NN closure development by using proxy–physics turbulence surrogates.
They used algebraic Reynolds stress models (ARSM) based turbulence surrogates, which could represent the non–linearity and bifurcation characteristics observed in different regimes of turbulent flow, to generate the parameter–to–solution maps. This allowed the authors to explore the potential of using DNN to model turbulence closures across regimes, without generating large amounts of high-fidelity data \cite{taghizadeh2020turbulence}. In the current paper, our primary objective is to quantify and compare model uncertainties within turbulence closure surrogates using data from proxy-physics surrogates. 

\subsection{NN Model of Turbulence Closure Trained using Proxy-Physics Surrogate}
%\begin{enumerate}
%        \begin{enumerate}

%            \begin{gather}
%                b_{ij} = G_1(s_{ij}) + G_2(s_{ik}r_{kj}-r_{ik}s_{kj}) + G_3(s_{ik}s_{kj} - \frac{1}{3}\delta_{ij}s_{mn}s_{nm}) + G_4(r_{ik}r_{kj} - \frac{1}{3} \delta_{ij}r_{mn}r_{nm}) \\
%                \frac{D b_{ij}}{Dt} = \frac{\partial b_{ij}}{\partial t} + U_{k} \frac{\partial b_{ij}}{\partial x_k} \approx 0 \\
%                (\eta_1 L_1^1)^2 G_1^3 - (2\eta_1 L_1^0 L_1^1)G_1^2 + \left[ (L_1^0)^2 + \eta_1 L_1^1 L_2 - \frac{2}{3} \eta_1(L_3)^2 + 2 \eta_2 (L_4)^2\right]G_1 - L_1^0L_2 = 0
%            \end{gather}
In this study, we use a three-term self–consistent, nonsingular, and fully explicit algebraic Reynolds stress model (EARSM) \cite{girimaji1996fully}, with the pressure–strain correlation model proposed by Speziale, Sarkar, and Gatski (SSG) \cite{speziale1991modelling}, to generate stress-strain relationship datasets for different normalized strain and rotation rates. 
The purpose of an ARSM is to capture the characteristics of a fluid flow for different values of non-dimensional variables effectively. The details of the closure model have been illustrated succinctly. The NN models analyzed in the paper have been developed using data generated by the equations listed below. 
The closure coefficients $G_1$, $G_2$, and $G_3$ are calculated as a function of $\eta_1$ and $\eta_2$. $\eta_1 = s_{ij}s_{ij} = S^2$ and $\eta_2 = r_{ij}r_{ij} = R^2$  are the scalar invariant of the strain and rotation–rate tensors.

\begin{equation}
    G_1 = 
    \begin{cases}
        -\frac{p}{3} + \left( -\frac{b}{2} + \sqrt{D} \right)^{1/3} + \left( -\frac{b}{2} - \sqrt{D} \right)^{1/3} & D > 0 \\
        -\frac{p}{3} + 2 \sqrt{-\frac{a}{3}} \cos(\frac{\theta}{3}) & D < 0, \ b<0 \\
        -\frac{p}{3} + 2 \sqrt{-\frac{a}{3}} \cos(\frac{\theta}{3} + \frac{2 \pi}{3}) & D < 0, \ b>0
    \end{cases}
    \label{eq:G1}
\end{equation}   
\begin{equation}
    G_2 = \frac{-L_4 G_1}{L_1^0 - \eta_1 L_1^1 G_1} \quad G_3 = \frac{2 L_3 G_1}{L_1^0 - \eta_1 L_1^1 G_1}
    \label{eq:G2-3}
\end{equation}

Where the calculation begins with defining the SSG-related constants $C_1^0$, $C_1^1$, $C_2$, $C_3$, and $C_4$ as 3.4, 1.8, 0.36, 1.25, and 0.4 respectively. 
One can then calculate $L_1^0$, $L_1^1$, $L_2$, $L_3$, and $L_4$.

\begin{align}
L_1^0 = \frac{C_1^0}{2} - 1, \quad L_1^1 = C_1^1 + 2, \quad L_2 = \frac{C_2}{2} - \frac{2}{3}, \quad L_3 = \frac{C_3}{2} - 1, \quad L_4 = \frac{C_4}{2} - 1
\end{align}
The intermediate variables $p$, $q$, and $r$ are then calculated. 
\begin{align}
\begin{split}
     p &= - \frac{2 L_1^0}{\eta_1 L_1^1} \quad r = - \frac{L_1^0 L_2}{(\eta_1 L_1^1)^2} \\
     q &= \frac{1}{(\eta_1 L_1^1)^2} \left[ (L_1^0)^2 + \eta_1 L_1^1 L_2 - \frac{2}{3} \eta_1 (L_3)^2 + 2\eta_2 (L_4)^2 \right]
\end{split}
\end{align}
Finally, the intermediate values used for determining $G_1$ are,
\begin{align}
\begin{split}
a = (q - \frac{p^2}{3}), &\quad b = \frac{1}{27} (2p^3 - 9pq + 27r) \\
D = \frac{b^2}{4} + \frac{a^3}{27}, &\quad \theta = \cos^{-1}\left(\frac{-b/2}{\sqrt{-a^3/27}} \right)
\end{split}
\end{align}

More details about the ARSM-SGG model used for generating the data for developing the NN can be found in Taghizadeh et al. \cite{taghizadeh2021turbulence}.

\subsection{\label{sec:level2}NN Model Architecture and Performance}
The machine learning objective of this paper is to train Neural Networks to fit the ARSM data with ($\eta_1$, $\eta_2$) as inputs and ($G_1$, $G_2$, $G_3$) as outputs.
To generate training, validation, and testing data, each input point is generated at random uniformly in the log base 10 space.
Generating training and test points in the log space ensures areas of high change are represented adequately in the training data as a majority of the change in ($G_1$, $G_2$, $G_3$) happens when the inputs are between zero and 10.
This paper uses 80,000 training data points for training, 40,000 points for validation, and a 700 by 700 grid (490,000 points) for testing.
As the true function of the inputs is known, a large testing dataset is generated to ensure the accuracy of each Network is tested in a robust manner that approximates the true prediction error.
After generating the data, each of the inputs and outputs are transformed to simplify the training and increase accuracy.
For the inputs, $\eta_1$ and $\eta_2$ are transformed by applying a natural logarithm and then using Sci-Kit Learn's standard scaler.
The outputs are transformed using only Sci-Kit Learn's standard scalar as all of the outputs are between $(-1,1)$.
Scaling in this way has the effect of shrinking the areas of low change in the ARSM and reducing the complexity of the function.
In practice, this transformation reduces the necessary model parameters from nearly 40,000 to just under 1,000. 
The loss of each method is reported in root mean squared error (RMSE) to promote the interpretability of the error.

The Neural Network in this paper has 4 total layers (1 input, 2 hidden, 1 output) with a hidden node size of 20 (963 learnable parameters).
Each layer uses a ReLU activation function except for the last where a linear transformation is used because the outputs need to have varying scales and the ability to be negative.
Each of the 40 ensemble members is trained with a weight regularization of $10^{-7}$, batch size of 64, a decaying learning rate (successive halving) starting at 0.001, and Mean Squared Error (MSE) loss.
The Monte-Carlo Dropout network is trained with a dropout percentage of 0.1\% on all layers except the output, a batch size of 256, a decaying learning rate (successive halving) starting at 0.001, and Mean Squared Error (MSE) loss.
The Stochastic Variational Inference network is trained with a batch size of 1024, a constant learning rate of 0.001, a Multivariate Gaussian prior over the NN parameters, a gamma distribution prior over the output noise, and a Multivariate Gaussian likelihood with diagonal covariance.
The Exact Gaussian Process in this paper uses a Squared Exponential (or RBF) Kernel, a constant zero mean, and assumes no output noise.
The Approximate Gaussian Process in this paper uses a Matern kernel with a $\nu$ of 2.5, 1,000 inducing points, a constant zero mean, and assumes no output noise.

\begin{figure}[t]
    \centering
    \includegraphics[scale=0.3]{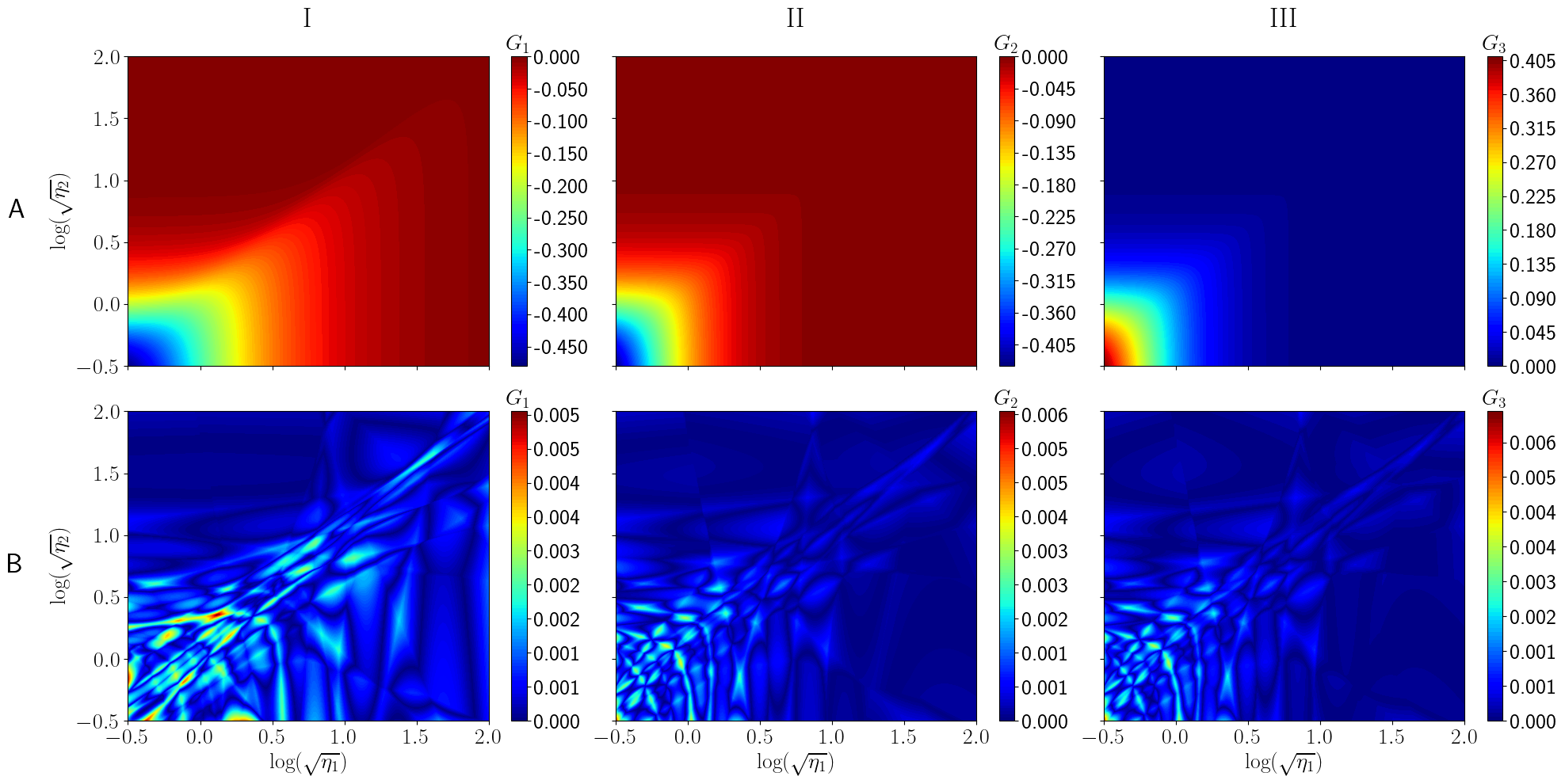}
    \caption{A plot of the true ($G_1$, $G_2$, $G_3$) from the ARSM (row A) and absolute error of Neural Network predictions (row B) with a test loss of $5.89\cdot 10^{-4}$ RMSE and $2.65\cdot 10^{-3}$ MAPE.}
    \label{fig:Deterministic_Preds}
\end{figure}

Fig. \ref{fig:Deterministic_Preds} row B shows the accuracy of a single deterministic network with identical training to the Ensemble networks.
As one can see, the prediction plots in row B are nearly indistinguishable from the reference plots in row A, which is reinforced by the RMSE of $5.89\cdot 10^{-4}$.
This shows the proposed network structure is sufficient to predict the turbulence closure parameters from the ARSM accurately.
In addition, comparing the results, the RMSE and MAPE testing errors of the proposed Neural Network are an order of magnitude lower than the results in \cite{taghizadeh2021turbulence}.
This likely stems from the transformations applied to the inputs and outputs of the problem, which enable small networks to achieve better accuracy by shrinking areas of low significance.

%%%%%%%%%%%%%%%%%%%%%%%%%%%%%%%%%%%%%%%%%%%%%%%%%
\section{UQ of NN Turbulence Closure Model}
\label{sec::nn-uq}

\subsection{Epistemic and Aleatoric Uncertainty}

In a machine learning problem, modelers try to learn a function from data by minimizing the loss of a model on a dataset.
\begin{equation}
    \theta^* = \argmin_{\theta} L(f_\theta, D)
    \label{eqn:ML_obj}
\end{equation}
Where $\theta$ represents the model parameters, $L$ is the loss function, $D$ is the dataset of input-output pairs, and $f_\theta$ is the model with parameters $\theta$.
The optimization problem in Eq. \ref{eqn:ML_obj} is often solved with Stochastic Gradient Descent (SGD) and is typically non-convex.
This means more than one minima exists in the loss function or, depending on the start, optimization of model parameters may converge to different configurations.
Some sets of model parameters may be algebraically equivalent, multiple parameter configurations create the same underlying function, or the parameters may underly entirely different functions.
Because one can not guarantee convergence to local minima, one can only conclude that a machine learning model learns a best estimate of the function generating the data.

The sources of uncertainty of for any phenomena can be broken into Aleatoric (Data) and Epistemic (Model) uncertainty.
Much of the uncertainty literature \cite{KIUREGHIAN2009105, gawlikowski2023survey} define Aleatoric uncertainty as the inherent randomness of a phenomena and Epistemic uncertainty as stemming from a lack of knowledge about the phenomena.
Aleatoric uncertainty is the uncertainty in the dataset or noise in the observations on which one wants to train a model.
Epistemic uncertainty is the uncertainty due to the lack of sufficient data, which in this paper is a Neural Network and Gaussian Process.
However, a more intuitive description of these uncertainties is irreducible (Aleatoric) and reducible (Epistemic) uncertainty.
Aleatoric uncertainty is irreducible because additional data points from the same noisy observer can not reduce the noise in the data.
Epistemic uncertainty is reducible because additional data points different from the current input space can reduce the number of potential functions that predict the data.

Epistemic uncertainty is the allowed variation of functions (model realizations) given training data and an assumed model structure.
Epistemic uncertainty can be separated into in-training (interpolation) and out-of-training (extrapolation) regimes of the input space.
In-training epistemic uncertainty is the uncertainty of a learned function within the regions contained in the training data.
This type of uncertainty can be thought of as 'holes' in the training data where the model is interpolating between data points.
Out-of-training epistemic uncertainty is the uncertainty of a learned function in regions not bounded by training data.
This type of uncertainty is unique because models that do not generalize well will exhibit high error and uncertainty as some function of distance from the training data.

\begin{figure}[h]
    \centering
    \includegraphics[scale=0.6]{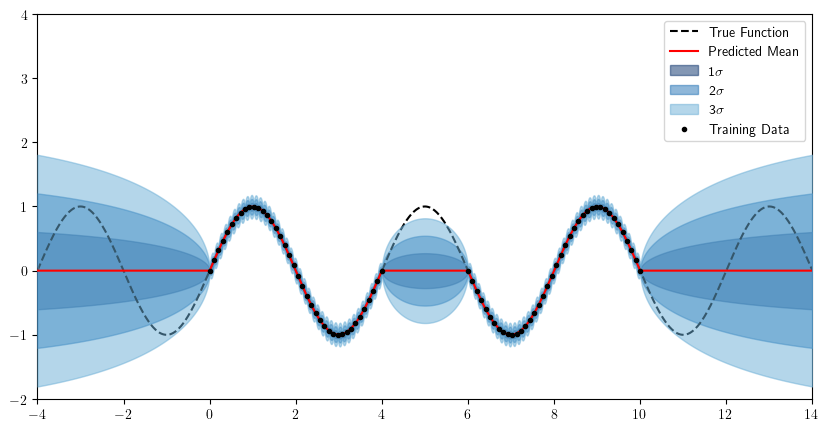}
    \caption{A figure showing Epistemic uncertainty in sparse regions of training data (inputs on the interval (4,6)) and regions not bounded by training data (intervals ($-\infty, 0$) and ($10, \infty$)).}
    \label{fig:Epistemic_ex}
\end{figure} 

Fig. \ref{fig:Epistemic_ex} shows the two different forms of Epistemic uncertainty in a model.
In the region of sparse training data on the interval (4,6) one can see the Epistemic uncertainty (in the form of standard deviations from the mean) is constrained because the model must pass through the points at 4 and 6.
As a result, the Epistemic uncertainty peaks at the midpoint of the gap in the data instead of going off unbounded.
Additionally, at the maxima and minima, the high curvature of the sin function introduces epistemic uncertainty even in regions with fairly dense training data.
Fig. \ref{fig:Epistemic_ex} also shows that without training data to inform the predictions, the epistemic uncertainty of a model may quickly encompass the output range.
This means a model that performs well on the training data may not learn the underlying function or what is often called "generalizing".
That is, it does not learn the periodic nature of the sin function.
Consequently, the epistemic uncertainty increases in the out-of-training region up to the output range as a function of some distance metric (in this case Euclidean).

Fig. \ref{fig:Epistemic_ex} highlights the importance of epistemic uncertainty estimates in the field of ML-based turbulence closure surrogates.
Although uncertainty quantification allows a modeler to imbue a degree of trust in predicted value, inaccurate uncertainty estimates can also be misleading.
Misleading uncertainty estimates are generally categorized into over-confident and under-confident estimates.
An over-confident uncertainty estimate is when the estimate is much smaller than the error for an input.
This will lead a modeler to believe a model prediction is trustworthy when, in reality, that prediction has high error.
An under-confident uncertainty estimate is when the estimate is much larger than the error of an input.
This will lead a modeler to believe a prediction is untrustworthy when the prediction is relatively accurate.
As one can already surmise, an ideal method of uncertainty quantification will give accurate uncertainty estimates in both in-training and out-of-training regimes.

\subsection{The Predictive and Posterior Distribution}

In statistics, there are two main categories of thought, Frequentist and Bayesian. 
Frequentist statistics focuses on the frequencies of repeated experiments or the ability to sample datasets from a source of truth.
Bayesian statistics focuses on the probability of an outcome given ones prior beliefs about the experiment and updating ones beliefs after observing data.
The Bayesian treatment is often the most commonly used because one usually only has a single data set to describe a phenomena and, in a regression setting, one may never observe exactly the same input twice.

As outlined above, the Bayesian approach models uncertainty by accounting for the probability of a the model parameters fitting the data.
One can then find the distribution of a modeled quantity given a dataset by accounting for all possible model estimates using the posterior predictive distribution.
\begin{equation}
    p(y|x,D) = \int_{\Theta} p(y|x,\theta)p(\theta | D) d\theta
    \label{eq:pred_dist}
\end{equation}
Eq. \ref{eq:pred_dist} can be read as the probability distribution of an output given an input and a dataset.
The right-hand side of Eq. \ref{eq:pred_dist} is a weighted average over all possible parameters $\theta$ of the probability of the output given the input and parameters.
More intuitively, the predictive distribution characterizes the possible values y can take on for all possible models that map x to y where models that better fit the dataset get more influence.
Eq. \ref{eq:pred_dist} is an expectation of $p(y|x,D)$ over $\theta$ and allows for the use of Monte-Carlo integration, also referred to as model averaging, to get an approximation of the predictive distribution.
\begin{equation}
    p(y|x,D) \approx \frac{1}{N} \sum_{\theta_i \in \Theta} p(y|x, \theta_i) \quad \theta_i \sim p(\theta |D) 
    \label{eq:MC-pred}
\end{equation}
Using Eq. \ref{eq:MC-pred} one can approximate the characteristics of $p(y|x,D)$ with a finite number of parameter samples from $P(\theta |D)$.
For instance, the mean and covariance in regression are,
\begin{equation}
    \hat{y} = \frac{1}{N} \sum_{\theta_i \in \Theta} f_{\theta_i}(x) \label{eq:mean}
\end{equation}
\begin{equation}
    \Sigma = \frac{1}{N-1} \sum_{\theta_i \in \Theta} (f_{\theta_i}(x) - \hat{y})^T(f_{\theta_i}(x) - \hat{y}) \label{eq:cov}
\end{equation}
In practice, it is very difficult to generate accurate samples $p(\theta |D)$ because its components are intractable to compute.
This intractability leads to the necessity methods to sample from $p(\theta |D)$, like Markov-Chain Monte-Carlo and Variational Inference.

$p(\theta |D)$ is what is known as the posterior distribution of $\theta$ and its order of conditioning on the data is not easy to evaluate.
However, one can express the posterior distribution in an evaluatable way using Baye's Rule.
\begin{equation}
    p(\theta |D) = \frac{p(D|\theta) p(\theta)}{p(D)} = \frac{p(D|\theta)p(\theta)}{\int p(D | \theta)P(\theta) d\theta}
    \label{eq:posterior}
\end{equation}
Thanks to Baye's rule, one can interpret the posterior distribution of $\theta$ as the multiplication of the probability of the parameters explaining the data and the probability of the parameters occurring.

The probability of the parameters explaining the data, $p(D|\theta)$ is often referred to as the likelihood of the data given some parameters $\theta$.
For Neural Networks, this likelihood can be related to the loss function of a Neural Network on the training data, or Neural Networks with higher accuracy have a higher probability of explaining the data (one can also show that minimizing MSE loss is the same as maximizing the Gaussian likelihood of the data).
The probability of the parameters $p(\theta)$ in Eq. \ref{eq:posterior} is referred to as the 'prior' distribution or how one incorporates prior knowledge into the distribution.
For example, using a Gaussian prior where all of the parameters are centered at zero is equivalent to regularization on the parameters of the network.
To make this link clearer, a Gaussian prior on the parameters centered at 0 makes parameters with a small magnitude more probable.
Similarly, regularization of the parameters of a network encourages their magnitude to be small, which is equivalent to making smaller magnitude parameters more probable.

The final term, and most difficult to evaluate, is the denominator in Eq. \ref{eq:posterior} and is the probability of the data over all possible parameters.
As one can already see, the integral for any NN is impossible to evaluate analytically over the entire parameter space.
However, this integral evaluates to a constant or only normalizes the posterior so it can remain a valid probability distribution.
Because it is a constant, researchers have devised ways in order to get around evaluating the integral and generate parameter samples from the posterior so they can be used to approximate the predictive distribution.

\section{Approximate Bayesian Inference Methods}
\label{sec::nn-bi}

\subsection{Deep Ensembles}

%\subsubsection{Overview}
In machine learning, an ensemble is created by combining the knowledge of multiple learners by taking a weighted sum of their outputs.
The proportion of the contribution can be different in the case of boosting, where learners are trained sequentially and to compensate for the error of the previous learners. 
Or the contribution can be the same, which facilitates the parallel training of the learners.
\begin{equation}
f_{\Theta}(x) = \frac{1}{N}\sum_{i=1}^N f_{\theta_i}(x) \quad \text{or} \quad \sum_{i=1}^N \alpha_i f_{\theta_i}(x) \ s.t. \ \sum \alpha_i = 1
\label{eq:Ensemble}
\end{equation}
The use of ensembles is especially useful when a model class has low bias and high variance, which allows for averaging to reduce the variance of predictions.
A more mathematically rigorous notion is the posterior distribution of a Deep Ensemble is approximated as the mixture of delta functions.
\begin{equation}
    P(\theta |D) = \frac{1}{N}\sum_{i=1}^N \delta(\theta - \theta_l)   
    \label{eq:ensemble_post}
\end{equation}
Eq. \ref{eq:ensemble_post} shows each ensemble member has an equal probability of occurring and the posterior has point densities at the parameters of each ensemble member.

Although a single model trained with maximum likelihood estimation (MLE) is frequentist, Gordon\cite{gordon2021} and Hoffman\cite{hoffmann2021deepensemblesbayesianperspective} show an Ensemble of deterministic networks is a form of approximate Bayesian Inference.
This equivalence is more apparent after linking the use of SGD on a loss function and finding Maximum A-Posteriori (MAP) estimates of the posterior.
Consider a typical Mean Squared Error Loss function of a Neural network with L2 regularization over an independent and identically distributed (IID) dataset $D$.
\begin{equation}
    \mathcal{L}(\theta,D) = \sum_{(x_i,y_i)\in D} ||y_i - f_{\theta}(x_i)||^2 + \lambda ||\theta||^2
    \label{eq:L2_loss}
\end{equation}
Where $f_\theta$ is the function represented by the NN with parameters $\theta$ and $\lambda$ is some importance on the regularization term.
Now consider the posterior probability of a particular $\theta$ with $y_i \sim \mathcal{N}(f_\theta(x_i), I)$ likelihood and $\theta \sim \mathcal{N}(0,\frac{1}{\lambda}I)$ prior.
\begin{equation}
    p(\theta |D) \propto \prod_{(x_i,y_i)\in D} \exp(-\frac{||y_i - f_\theta(x_i)||^2}{2}) \exp(-\frac{\lambda||\theta||^2}{2})
    \label{eq:gauss_post}
\end{equation}
Where the likelihood can be expressed as a repeated product over the entire dataset from the IID assumption.
One can see minimizing the loss in Eq. \ref{eq:L2_loss} is equivalent to minimizing the negative logarithm of the posterior in Eq. \ref{eq:gauss_post}.
This shows a deterministic network trained with SGD on Eq. \ref{eq:L2_loss} is equivalent to a MAP sample of $\theta$ from its corresponding posterior distribution.

The main difficulty of Deep Ensembles is maintaining functional diversity among the different members.
In some problems, (\cite{Deep-Ensembles}, \cite{fort2020deep}) random initialization of parameters and shuffling of data can be sufficient to promote functional diversity among members of a Deep Ensemble.
However, this is not generally true as Gal \cite{galDissertation} illustrates a simple example where the functional diversity of a Deep Ensemble collapses.
There exists a wide variety of ways to combat the functional collapse of a Deep Ensemble, which mostly stem from improving the approximation of the Bayesian inference.
For example, Pearce \cite{pearce2018bayesianinferenceanchoredensembles} and Casaprima \cite{10.1115/1.4069339} introduce Anchored Ensembles, which 'anchors' a member of the ensemble around a sample from the prior.
Another promising approach is incorporating what is called a functional prior \cite{dangelo2023repulsivedeepensemblesbayesian, ghorbanian2024empoweringbayesianneuralnetworks, tran2022needgoodfunctionalprior} into the posterior $p(\theta |D)$.
A functional prior is much different than a prior over the parameters and addresses the issue of parameters being identical in function space but very different in the parameter space.

The disadvantage of Deep Ensembles is the linear scaling of training, compute resources and storage.
First, Deep Ensembles require training several Neural Networks that scale linearly with time if not trained in parallel.
Many ways exist to reduce the training time to acquire members of the ensemble, such as large cyclic learning rates, but at the expense of functional diversity.
Second, Deep Ensembles linearly scale in computational resources for storage and inference.
For example, when using GPUs for inference, one will have to store a number of networks on the GPU's dedicated memory for inference or distribute members across multiple GPUs for faster inference.
Lastly, each ensemble member is unique, which linearly increases the amount of data to store an Ensemble for later use.

\subsection{Monte-Carlo Dropout (MC-Dropout)}
%\subsubsection{Overview}
Monte-Carlo Dropout is widely used as a regularization technique for large networks to improve generalization.
Monte Carlo Dropout consists of adding a diagonal matrix after applying the non-linearity, where the entries are generated by a Bernoulli distribution $z_{ii} \sim Bernoulli(p)$.
These Dropout matrices are then generated for each prediction of the NN which 'drops' p \% of the elements in an output vector (nodes) for a given layer of a neural network.
In the limit of an infinitely wide NN, \cite{gal2016dropout} shows any network structure with Monte-Carlo Dropout is equivalent to performing Variational Inference of a Deep Gaussian Process.
Where a Deep Gaussian Process is the composition of multiple Gaussian Processes, which are introduced later as a way of quantifying uncertainty.
Although not infinite, a Neural Network trained with Monte-Carlo Dropout is a collection of parameters where any arbitrary subset is a function that approximates the data.
This means the training of the network creates the posterior in the form of a Neural Network and one can generate samples from the posterior by randomly dropping out p\% nodes.

The benefit of MC-Dropout is that it circumvents the necessity of storing multiple unique networks, which reduces training time and storage.
However, the benefits come at the cost of introducing new hyper-parameters, a tendency toward single-modal approximations, and reducing model expressiveness.
First, MC-Dropout requires tuning of the p parameter for each layer depending on each dataset and the amount of data.
This means each layer of the network potentially requires a different dropout rate p with no steadfast rules for finding the optimal values.
However, \cite{gal2017concrete} propose a way to optimize p for each layer using gradient methods using Concrete Dropout, a variation on Monte-Carlo Dropout.
Second, each sample from the MC-Dropout posterior tends to be correlated with one another, affecting function diversity and uncertainty estimates.
Lastly, the MC-Dropout training reduces the complexity of the proposed Neural Network functions through a reduction in parameters.
This can reduce overfitting but at the cost of introducing more variance in each of the sampled functions from the MC-Dropout posterior.
However, methods such as Drop-Connect \cite{dconnect-wan13} attempt to reduce the imposed variance by zeroing individual weights instead of entire nodes.

\subsection{Stochastic Variational Inference (SVI)}

%\subsubsection{Overview}
Stochastic Variational inference is yet another way of characterizing a posterior distribution over the weights of a Neural Network given some data.
In general, SVI is an extension of standard Variational Inference where the gradients are estimated from subsets of the data \cite{hoffman2013stochastic}.
The key part of this technique is defining a variational distribution which greatly simplifies some of the computation for inference.
The optimization algorithm then maximizes the Evidence Lower Bound using SGD, which minimizes the KL divergence between the variational distribution and true posterior.
\begin{equation}
\argmin_{q} D_{KL}(q(\theta) || p(\theta || D)) = \argmax_q \ E_{q(\theta)}\left[\log(p(\theta | D)\right] - D_{KL}(q(\theta) || p(\theta))
\label{eq:ELBO}
\end{equation}
In simpler terms, the objective of Eq. \ref{eq:ELBO} is to make a simpler distribution match a much more complicated true posterior.
However, it is important to draw attention to the expectation term in Eq. \ref{eq:ELBO} because this value must be estimated with samples from the variational distribution $q(\theta)$.
This induces extra noise on top of the noise present in estimating the gradient of the variational parameters of $q(\theta)$, increasing the difficulty of the optimization process.
In this paper, a Multivariate Gaussian variational distribution with a full-rank covariance matrix is employed to estimate the posterior over the Network parameters.
This results in an NN posterior that uses a Multivariate Gaussian distribution over the parameters to model a single mode of the posterior or NN function.
Samples from this simpler distribution can then be easily generated to quantify the uncertainty of predictive distribution, where each sample is a set of weights and biases of the neural network.

The benefit of using SVI is twofold, its ability to scale to larger networks and its potential to reduce storage.
As a result of the objective in Eq. \ref{eq:ELBO}, SVI uses SGD to optimize the parameters of its variational distribution which scales well with large networks and can compensate for its lack of expressiveness.
In addition, compared to an Ensemble, the use of SVI has the potential to decrease the storage of parameters (matrix of mean values, covariance matrix), depending on the assumption of the covariance matrix.
For some problems a diagonal covariance matrix only requires the doubling of storage, whereas other variational distribution assumptions may require a large number of ensemble members to produce a storage improvement over a Deep Ensemble.

The disadvantages of SVI are lack of functional diversity, difficult optimization objective, and the limitations of variational distribution.
As mentioned previously, SVI characterizes a tractable distribution over the a neural networks which reduces its expressiveness. 
This can lead to low-quality uncertainty estimates as the characterization of multiple functions may be necessary.
However, this effect can be mitigated by training an ensemble of SVI networks to characterize multiple functions.
Another limitation to SVI is the ELBO optimization objective is a more abstract and difficult quantity to calculate.
For instance, the expectation of the log-likelihood must be estimate using Monte-Carlo, introducing noise into the optimization objective and degrading accuracy.
The final limitation of SVI is the proposed variational distribution over the parameters of a Neural Network.
This means assuming some structure on the interaction between the parameters in the Neural Network may reduce the complexity of the functions a network trained with SVI can represent.
Consequently, the accuracy of a network trained with SVI may be significantly worse than that of a deterministic network with the same structure.

\subsection{Gaussian Processes}

A Gaussian Process is a Non-Neural Network-based way of quantifying the Epistemic uncertainty of a model.
Rasmussen\cite{RasmussenGP} defines a Gaussian Process (GP) as a collection of random variables where any finite number of them have a joint Gaussian distribution or Multivariate Normal Distribution.
More generally, a GP is a stochastic process that can be assumed to be noiseless or noisy depending on the data.
For a noiseless dataset, one assumes the true output is observed in the data such that $f(x) = y$.
For a noisy dataset, one assumes the observed outputs are centered on the true value and perturbed by some random noise $f(x) = y + \epsilon$.
Where $\epsilon$ is Gaussian noise centered at zero with variance $\sigma_n^2$ or $\epsilon \sim N(0,\sigma_n^2)$.
Using the noisy formulation with N training inputs $x$, N training outputs $y$, D test inputs $x^*$, and D unknown test outputs $f(x^*)$, the foundational assumption of a GP is that each output follows a Multivariate Gaussian Distribution.
\begin{equation}
    \begin{bmatrix} y \\ f(x^*)\end{bmatrix} \sim N\left( \begin{bmatrix} m(x) \\ m(x^*) \end{bmatrix}, \begin{bmatrix}
        K_{xx} + I\sigma^2_n & K_{xx^*} \\ K^T_{xx^*} & K_{x^*x^*}
    \end{bmatrix} \right)
    \label{eq:joint_gp}
\end{equation}
Where $K_{xx}$ denotes an NxN symmetric covariance matrix between the training inputs, $K_{xx^*}$ denotes an NxD covariance matrix between the training and testing inputs, and $K_{x^*x^*}$ is the DxD symmetric covariance matrix between the testing inputs.

In probability, one usually refers to Eq. \ref{eq:joint_gp} as the joint probability distribution over the training and testing outputs $p(f(x^*), y | x, x^*)$.
One can also notice Eq. \ref{eq:joint_gp} is very similar to the predictive distribution in Eq. \ref{eq:pred_dist}, except $y$ is assumed to be random and we have neglected the parameters of the model $\theta$.
However, the Bayesian formulation assumes the model's parameters are random and not the data.
As a result, one can assume $y$ is known and constant, which allows the GP to be conditioned on the training outputs through the Gaussian Conditioning property (derivation in Rasmussen \cite{RasmussenGP}).
This then leads to a closed-form solution for the predictive distribution over $f(x^*)$ necessary for uncertainty quantification, where $\Sigma$ is $K_{xx} + I\sigma_n^2$.
\begin{equation}
    p(f(x^*) | x, x^*, y) = N\left(m(x^*) + K^T_{xx^*}\Sigma^{-1}(y-m(x^*)), K_{x^*x^*} - K^T_{xx^*}\Sigma^{-1}K_{xx^*}\right)
    \label{eq:gp_pred}
\end{equation}
Eq. \ref{eq:gp_pred} gives the mean prediction for the testing points and the standard deviation of the predictions as the square root of the diagonal of the covariance, similar to the NN methods.

In Eq. \ref{eq:gp_pred}, the mean $m(x)$ and covariance function $k(x,x')$ entirely specify the GP.
Most commonly, one assumes the mean to be the zero function ($m(x) = 0)$ and the covariance to be the squared exponential (SE) ($k(x_i,x_j) = \sigma_1 exp(-\frac{1}{2\sigma_2^2} ||x_i-x_j||^2)$).
A more in-depth explanation of mean and covariance selection can be found in Rasmussen \cite{RasmussenGP}.
The mean and covariance function of the GP reveals the true parameters of the predictor, which in the Bayesian formulation we assume to be random.
With zero mean and SE covariance, the parameters of a GP are $\theta = \{\sigma_n, \sigma_1, \sigma_2\}$.
To infer the parameters $\theta$ from the GP, we use the posterior distribution over the parameters using only the training data as follows:
\begin{equation}
    p(\theta | x, y) \propto N(y | x, \theta)p(\theta) \,,
    \label{eq:gp_posterior}
\end{equation}
where $N(y | x, \theta)$ is the Gaussian likelihood of the training data and $p(\theta)$ is some prior over the parameters of the covariance function.
One can see Eq. \ref{eq:gp_posterior} looks eerily similar to Eq. \ref{eq:posterior}, which means one can sample the parameters of the GP using most methods detailed in this paper.
Most commonly, Eq. \ref{eq:gp_posterior} is optimized over the parameters using SGD to produce a MAP estimate of the parameters by using an uninformative prior.

The main benefit of using a GP for Epistemic uncertainty quantification is the robust uncertainty estimates, as the GP integrates over all possible functions of the training data with mean $m(x)$ and covariance $k(x_i,x_j)$.
Although this is not obvious at first, Eq. \ref{eq:gp_pred} gives a closed-form solution to sample entire functions that explain the data.
As a result, the mean and covariance of predictions in Eq. \ref{eq:gp_pred} are not approximations and one can sample functions from the distribution.
The main drawback is the storage and inversion of the data covariance matrix in Eq. \ref{eq:gp_pred}.
This prevents GP's from being used in contexts with a significant amount of data ($n > 10,000$) because of the O($n^3$) time and O($n^2$) space complexity of matrix inversion.
In addition, the matrix inversion of the NxN covariance matrix of training inputs must be done multiple times when sampling the covariance function parameters, which compounds the issue.
This highlights the trade-off of a GP, having very few hyperparameters, compared to a NN, at the cost of inverting and storing large covariance matrices.

Rasmussen \cite{RasmussenGP} gives multiple ways of circumventing the large matrix inversion by using Low-rank or other approximations to the GP.
In addition to an Exact GP, this paper uses a Variational inducing point approximate GP from Titsias\cite{approxgp-titsias09a} implemented in GPytorch \cite{gpytorch-gardner2021}.
The principle of the Variational inducing point approximation is to use a subset of the training inputs to achieve satisfactory accuracy while avoiding large matrix inversions.
The Variational inducing point approximation is used in this paper because it represents a potentially viable method for modelers with large datasets, such as turbulence closure.

\section{Results}
\label{sec::nn-results}

\subsection{In-Training UQ Results}

This section shows the performance and UQ accuracy of Deep Ensembles, Monte-Carlo Dropout, Stochastic Variational Inference, Exact Gaussian Process, and Approximate Gaussian Processes using the full input space as training data.

\begin{figure}[h]
    \centering
    \includegraphics[scale=0.25]{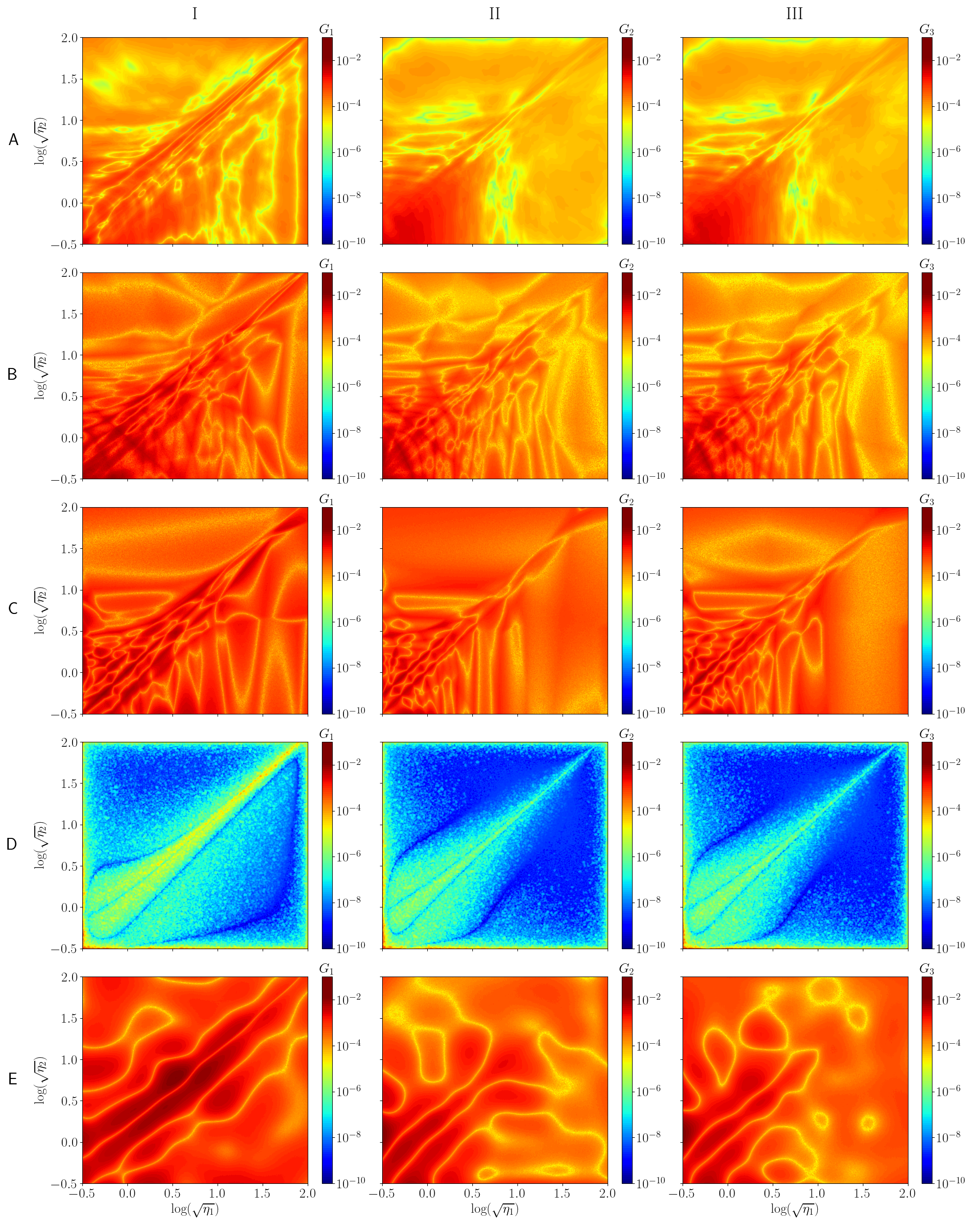}
    \caption{A plot of the absolute error between predictions and true values of ($G_1$, $G_2$, $G_3$) on the full input space for a Deep Ensembles (row A, RMSE $4.59\cdot 10^{-4}$), Monte-Carlo Dropout (row B, RMSE $8.45\cdot 10^{-4}$), Stochastic Variational Inference (row C, RMSE $8.76\cdot 10^{-4}$), Exact Gaussian Process (row D, RMSE $2.14\cdot 10^{-5}$) and Approximate Gaussian Process (row E, RMSE $1.68 \cdot 10^{-3}$).}
    \label{fig:Error_Full}
\end{figure}

Fig. \ref{fig:Error_Full} shows the accuracy of each of the different methods in terms of Absolute Error.
One can notice the Exact Gaussian Process has significantly better accuracy than the other methods.
This is because the Exact Gaussian Process has access to the entirety of the training data through the covariance matrix.
However, this is not without downside as the amount of data in this study required significant computational resources for training.
Conversely, the Neural Network-based models rely entirely on the relationships drawn from learning the training dataset.
The perfomance of the Deep Ensemble is the best of the NN methods but nowhere near the Exact Gaussian Process.
Fig. \ref{fig:Error_Full} also shows, as expected, the use of Monte-Carlo Dropout, Stochastic Variational Inference, and approximate Gaussian Process methods decreases the overall accuracy of each of the models over the deterministic results in Fig. \ref{fig:Deterministic_Preds}.

\begin{figure}[h]
    \centering
    \includegraphics[scale=0.25]{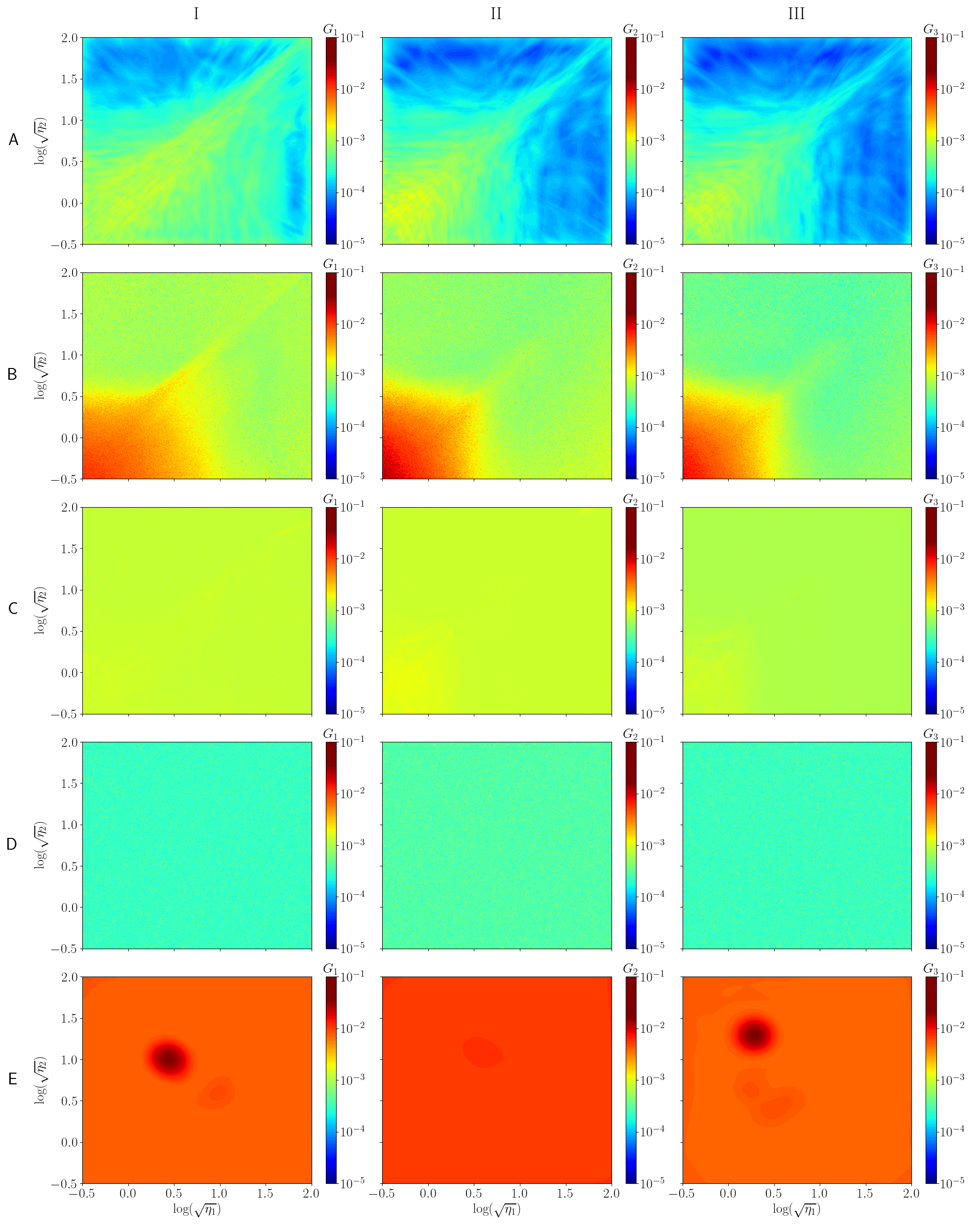}
    \caption{A plot of the standard deviation (square root of the diagonal of the covariance matrix) of the predictions in the full input space for a Deep Ensemble (row A), Monte-Carlo Dropout (row B), Stochastic Variational Inference (row C), Exact Gaussian Process (row D), and Approximate Gaussian Process (row E).}
    \label{fig:STD_Full}
\end{figure}

Fig. \ref{fig:STD_Full} shows the uncertainty in terms of standard deviations for all of the methods.
The Exact Gaussian Process and Deep Ensembles have the first and second best uncertainties respectively throughout the input space.
Most interestingly, the Deep Ensembles seem to exhibit lower uncertainties than the Exact GP in the outer regions where the outputs are essentially zero.
This shows an advantage of Deep Ensembles over the Exact GP because the uncertainty of the Exact GP with an SE kernel is dependent on the distance between a test point and the training points.
Whereas the uncertainty of the Deep Ensemble in regions is due to the differing output predictions among the different members of the Ensemble.
The Approximate GP, MC-Dropout, SVI NN's exhibit the highest uncertainties out of the four methods and show the difficulty of each method to model the function from the data.
In the case of SVI and MC-Dropout, the high uncertainties are likely due to an increase in gradient noise during their training.
For the Approximate Gaussian Process, the high uncertainty is because it is likely difficult to characterize the entire input space using only 1,000 inducing points.

\subsection{Out-of-Training UQ}

\begin{figure}[h]
  \centering
  \includegraphics[scale=0.5]{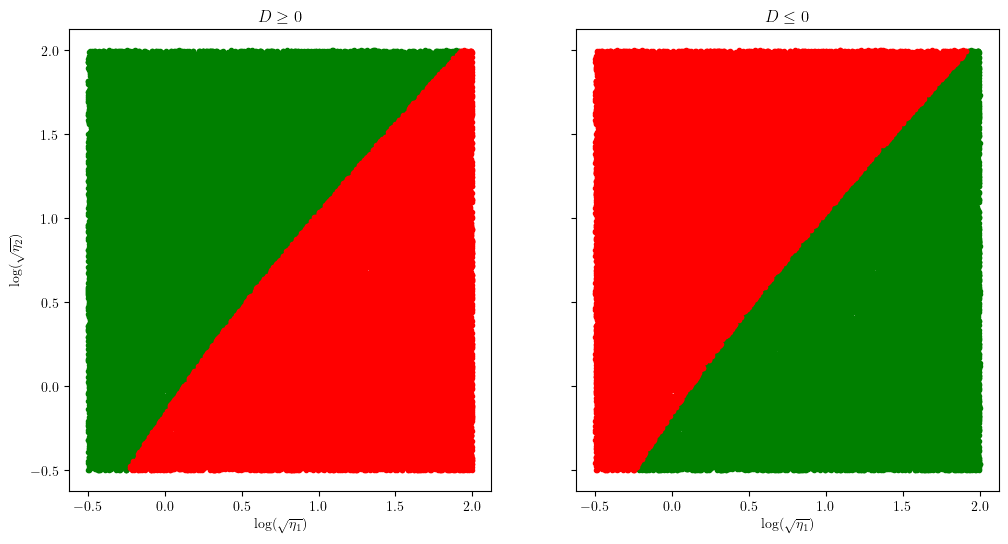}
  \caption{A plot of the training (green) and out-of-training (red) input regions for the two cases where $(\eta_1, \eta_2)$ satisfy $D \ge 0$ (left) and $D \le 0$ (right).}
  \label{fig:out_of_train_regions}
\end{figure}

Fig. \ref{fig:out_of_train_regions} shows the training input regions for the cases in Eq. \ref{eq:G1} where $D \ge 0$ and $D \le 0$.
The intent of this study is to determine the effect of not using the entire input space on the uncertainty quantification of each of the methods.
Most importantly, this section studies the robustness of a methods uncertainty estimates relative to the error in the out-of-training input regime.

\begin{figure}[h]
    \centering
    \includegraphics[scale=0.25]{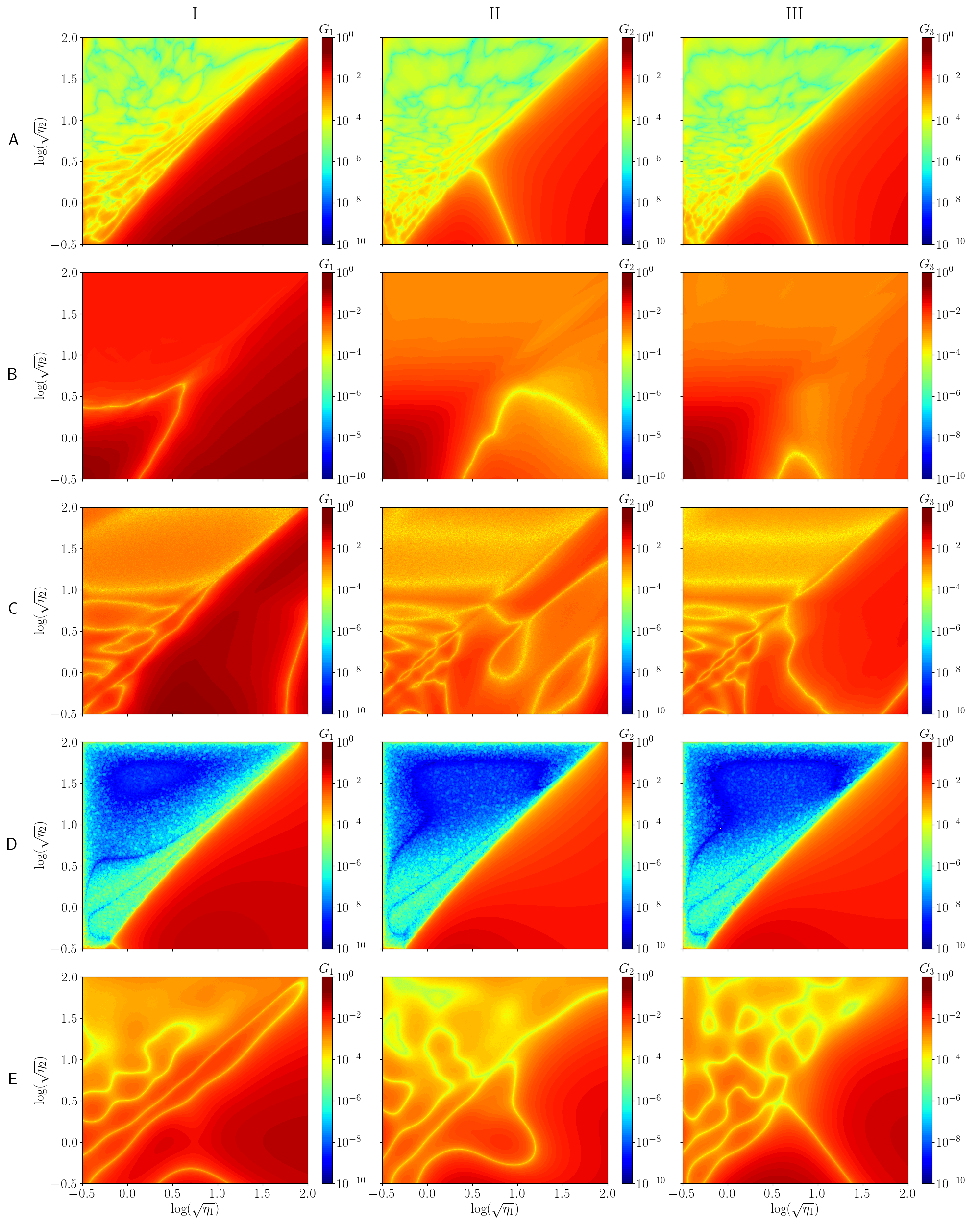}
    \caption{A plot of the absolute error between predictions and true values of ($G_1$, $G_2$, $G_3$) on the $D\ge0$ case for a Deep Ensembles (row A, RMSE $3.75\cdot 10^{-2}$), Monte-Carlo Dropout (row B, RMSE $4.26\cdot10^{-2}$), Stochastic Variational Inference (row C, RMSE $2.81\cdot10^{-2}$), Exact Gaussian Process (row D, RMSE $1.77\cdot 10^{-2}$), and Approximate Gaussian Process (row E, RMSE $2.05\cdot 10^{-2}$).}
    \label{fig:DGT0_Pred_Error}
\end{figure}

Fig. \ref{fig:DGT0_Pred_Error} shows the absolute error of each of the methods using training data only from inputs where $D\ge0$.
As would be expected, all of the methods perform poorly in the input region not included in the training data.
This suggests generalization in turbulence closure problems may not be achievable on more complex problems.
Fig. \ref{fig:DGT0_Pred_Error} shows the Exact GP has the best error followed by the Approximate GP, SVI, Deep Ensembles and MC-Dropout.
Interestingly, DE performs better than the other NN-based methods in the training input space, but the SVI network performs better overall.
The error discrepency compared to the full data may be due to the ability of SVI to model the the function landscape around a specific function and reduce the overall error.
However, this may also simply be an artifact of a better initialization of the SVI network's variational distribution.

\begin{figure}[h]
    \centering
    \includegraphics[scale=0.25]{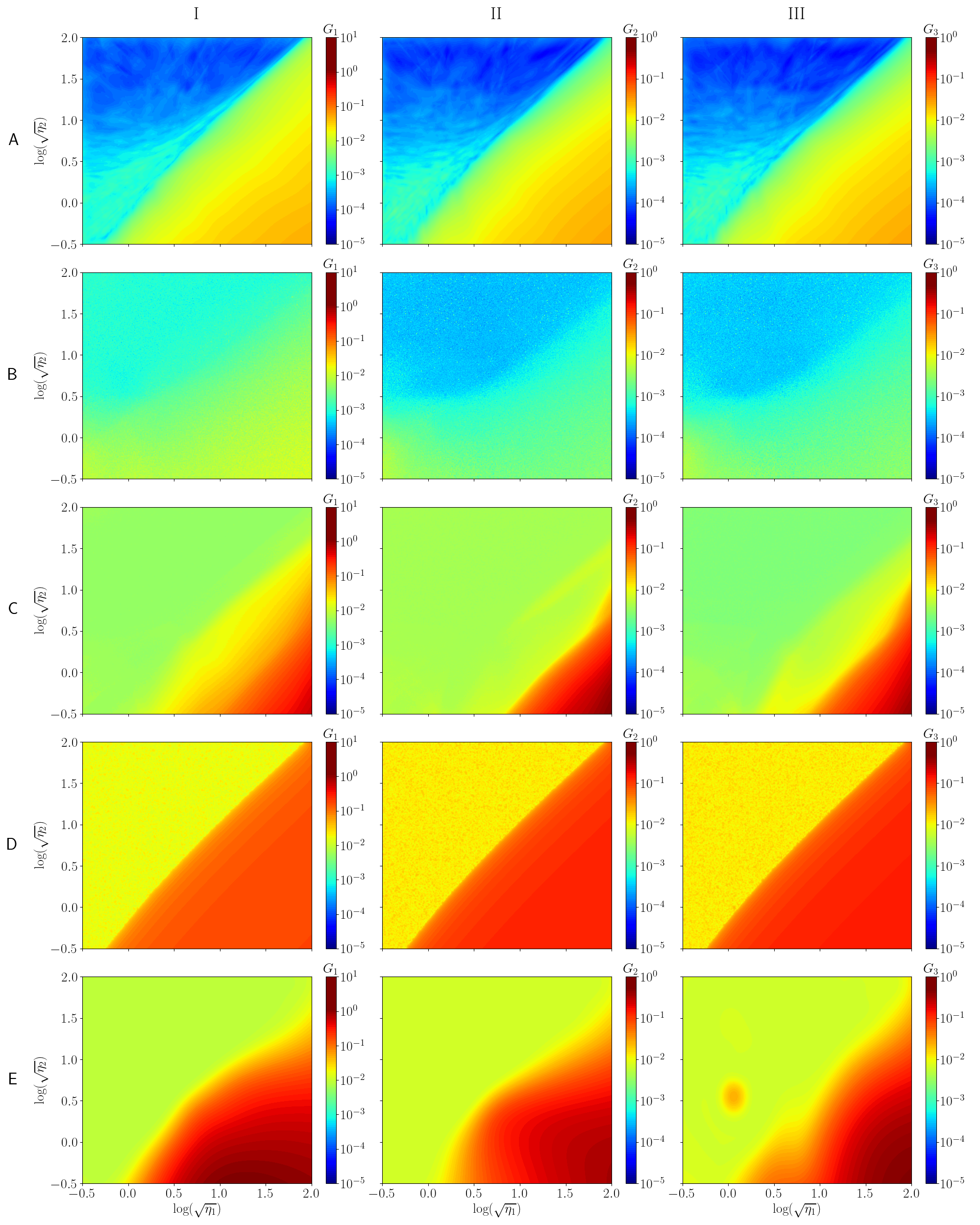}
    \caption{A plot of the standard deviation (square root of the diagonal of the covariance matrix) of the predictions in the $D\ge0$ case for a Deep Ensemble (row A), Monte-Carlo Dropout (row B), Stochastic Variational Inference (row C), Exact Gaussian Process (row D), and Approximate Gaussian Process (row E).}
    \label{fig:DGT0_Pred_STD}
\end{figure}

Fig. \ref{fig:DGT0_Pred_STD} show the epistemic uncertainty of each ML method on the out-of-training case where only inputs from the $D\ge0$ are included.
Looking at the overall result, one can see the uncertainty of each method is increasing in the out-of-training input region and trending with the error.
Comparing with Fig. \ref{fig:Error_Full}, one can see that Deep Ensembles and SVI are overconfident in the out-of-training region.
This is a similar result to Geneva \cite{Geneva_2020} where their Bayesian NN showed signs of over-confidence in RANS simulations.
In contrast, both the GP methods give more robust uncertainty estimates in the out-of-training region.
However, the Approximate GP shows signs of underconfidence in the $(2.0, -0.5)$ neightborhood of the input space.
In the Exact GP plots, one can see the relation of uncertainty with distance to the training data as there is visible peaks of uncertainty within the training region.
This may indicate overfitting of the Exact GP because the outputs are fairly smooth in the space.

\begin{figure}[h]
    \centering
    \includegraphics[scale=0.25]{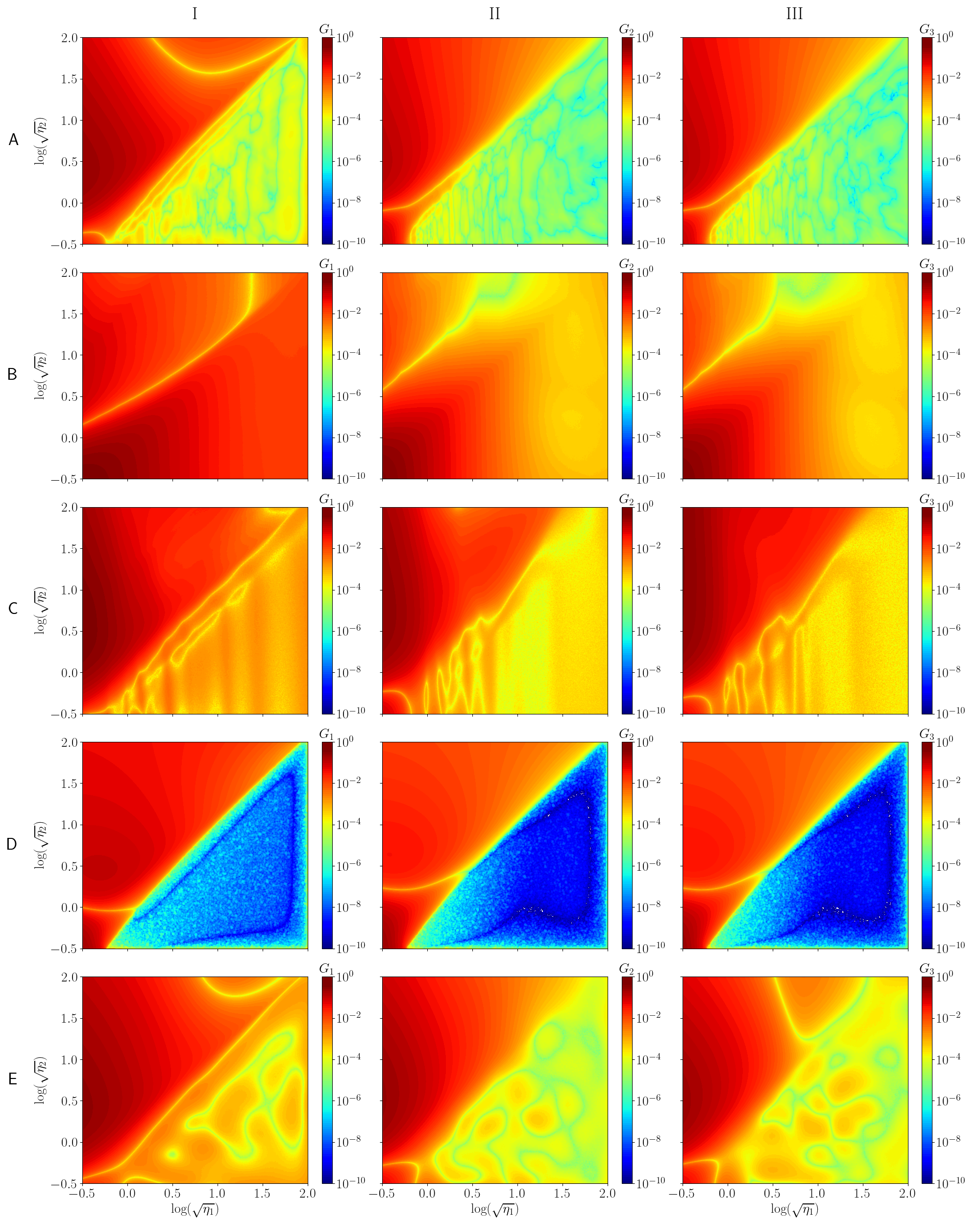}
    \caption{A plot of the absolute error between predictions and true values of ($G_1$, $G_2$, $G_3$) on the $D\le0$ case for a Deep Ensembles (row A, RMSE $4.14\cdot10^{-2}$), Monte-Carlo Dropout (row B, $5.89\cdot10^{-2}$), Stochastic Variational Inference (row C RMSE, $8.23\cdot10^{-2}$), Exact Gaussian Process (row D RMSE, $2.76\cdot10^{-2}$), and Approximate Gaussian Process (row E RMSE, $5.90\cdot10^{-2}$).}
    \label{fig:DLT0_Pred_Error}
\end{figure}

Fig. \ref{fig:DLT0_Pred_Error} shows the error of each of the ML methods where each is trained only on inputs where $D \le 0$.
Like the $D\ge0$ case, each of the methods performs significantly worse in the out-of-training input region.
This is more evidence that generalization of methods may be very difficult because error increases sharply when transitioning to the out-of-training region.
One can see the performance of the Exact GP is the best followed by Deep Ensembles, MC-Dropout, Approximate GP, and SVI.
The significant difference in error between the DE and SVI networks reinforces the error results in the $D \ge 0$ case was likely a result of initialization.
The most interesting result is the similarity in the plots of DE, SVI, and Approximate GP in the out-of training region.
The results show the highest error is not the point furthest from the training data but at the input $(-0.5, 0.5)$.

\begin{figure}[h]
    \centering
    \includegraphics[scale=0.25]{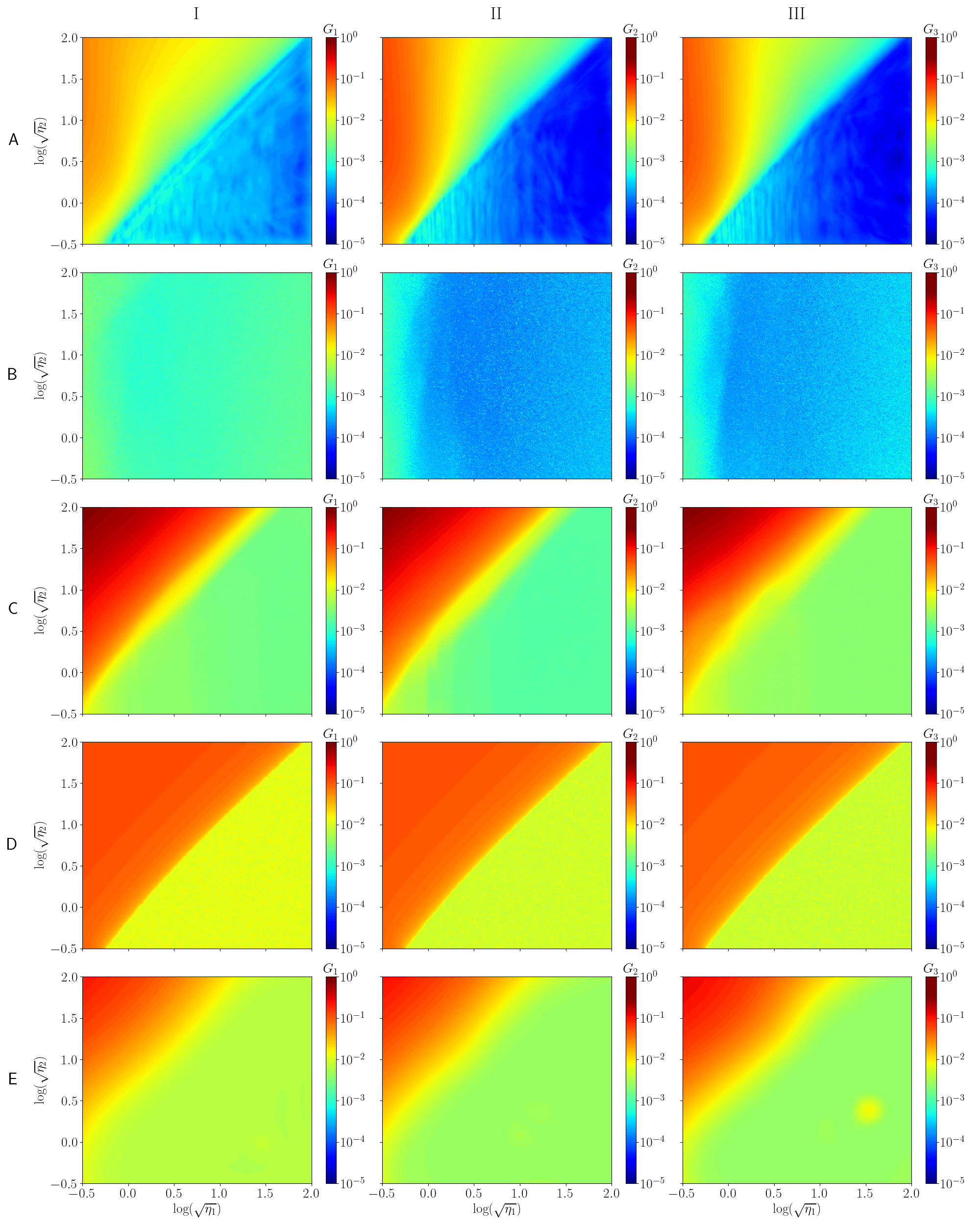}
    \caption{A plot of the standard deviation (square root of the diagonal of the covariance matrix) of the predictions in the $D\le0$ case for a Deep Ensemble (row A), Monte-Carlo Dropout (row B), Stochastic Variational Inference (row C), Gaussian Process 1 (row D), and Gaussian Process 2 (row E).}
    \label{fig:DLT0_Pred_STD}
\end{figure}

Fig. \ref{fig:DLT0_Pred_STD} shows the epistemic uncertainty of each ML method in the $D \le 0$ case.
One can see the Exact GP models the uncertainty the best followed by the DE.
Fig. \ref{fig:DLT0_Pred_STD} also shows the uncertainty of all methods besides the DE vary simply as a function of distance from the training data.
However, Fig. \ref{fig:DLT0_Pred_Error} shows the error does not follow this trend where the area of highest error is at $(-0.5, 0.5)$.
This is a promising result for the Deep Ensemble because it models the trend in uncertainty better than the rest.
However, the Deep Ensemble is not without flaws as it is over-confident on the border of the training data.

\section{Discussion}
\label{sec::nn-disc}

Each of the methods have their accuracy measured based on Mean Absolute Error (MAE) and Root Mean Squared Error (RMSE) detailed in Appendix \ref{append:UQ_Perf}.
The UQ estimates of each method are evaluated based on Miscalibration Area (MisCal), Calibration Error (CE), Sharpness (Sha), Coefficient of Variation ($C_v$), and Negative Log-Likelihood (NLL) all of which are detailed in Appendix \ref{append:UQ_Perf}.

% D GT 0 table
\begin{table}[H]
    \centering
    \setlength{\tabcolsep}{5pt}
    \renewcommand{\arraystretch}{1.3}
    \caption{A table of performance and uncertainty metrics for inputs $D \ge 0$ for Deep Ensembles (DE), Monte-Carlo Dropout (MCD), Stochastic Variational Inference (SVI), Gaussian Processes (EGP) and Approximate Gaussian Process (AGP).}
    \begin{tabular}{c|cccccc} 
        Method & MAE & RMSE & MisCal & Sha & $C_v$ & NLL \\ \hline \hline
        DE & $1.600\cdot 10^{-2}$ & $3.758 \cdot 10^{-2}$ & 0.138 & $2.357\cdot 10^{-4}$ & 2.327 & $-5.886\cdot10^{6}$ \\ 
        MCD & $2.217\cdot10^{-2}$ & $4.267\cdot10^{-2}$ & 0.485 & $1.138\cdot10^{-5}$ & 1.736 & $1.174\cdot10^{9}$ \\ 
        SVI & $1.269\cdot10^{-2}$ & $2.813\cdot10^{-2}$ & 0.111 & $6.311\cdot10^{-3}$ & 4.401 & -$2.756\cdot10^{6}$ \\ 
        EGP & $1.016\cdot10^{-2}$ & $1.772\cdot10^{-2}$ & 0.363 & $1.242\cdot10^{-2}$ & 1.549 & -$3.586\cdot10^{6}$ \\ 
        AGP & $1.130\cdot10^{-2}$ & $2.051\cdot10^{-2}$ & 0.285 & $1.112\cdot10^{-1}$ & 2.957 & -$4.125\cdot10^{6}$ \\ \hline \hline
    \end{tabular}
    \label{tab:dgt0_metrics}
\end{table}

Table \ref{tab:dgt0_metrics} shows the various performance and uncertainty metrics for each method in the $D \ge 0$ case.
Upon inspection, the error of each of the methods is significantly higher in the out-of-training case, which is consistent with results from original tests in Taghizadeh\cite{taghizadeh2020turbulence}.
Table \ref{tab:dgt0_metrics} shows quantitatively the GPs perform better than the NN methods.
Of the Neural Network methods, SVI out-performs the Deep Ensemble, but the gap is much less in the MAE metric compared to the RMSE.
This means DE performs well overall but there are specific regions with much higher error than SVI.
Looking at the UQ metrics, Miscalibration Area metric favors the NN methods.
This means the empirical distribution of the NN methods resemble a Gaussian distribution more than the Gaussian Process, an interesting result.
One can also see in the Sharpness metric the GP methods have a higher variance in their uncertainty indicating they are likely underconfident in regions with low error.
Overall, all methods suffer from over or under-confidence in certain regions of the input space.
This is reinforced by the results from Geneva \cite{geneva2019quantifying} and Zhang \cite{Zhang2015MachineLM}, where in the in-training case their Bayesian NNs suffered from over and under-confidence.
Looking at the Negative Log-Likelihood, an overall performance metric, the Deep Ensemble out-performs the other methods.
This indicates the Deep Ensemble allocates the highest probability to each true value on average compared to the other methods.
Where as a method such as MC-Dropout allocates very low uncertainty to some true outputs which is generally unacceptable.

% D LT 0 table
\begin{table}[H]
    \centering
    \setlength{\tabcolsep}{5pt}
    \renewcommand{\arraystretch}{1.3}
    \caption{A table of performance and uncertainty metrics for inputs $D \le 0$ for Deep Ensembles (DE), Monte-Carlo Dropout (MCD), Stochastic Variational Inference (SVI), Gaussian Processes (GP), and Approximate Gaussian Process (AGP).}
    \begin{tabular}{c|cccccc} 
        Method & MAE & RMSE & MisCal & Sha & $C_v$ & NLL \\ \hline \hline
        DE & $1.904\cdot10^{-2}$ & $4.141\cdot10^{-2}$ & 0.116 & $4.683\cdot10^{-4}$ & 2.858 & -$5.508\cdot10^{6}$ \\ 
        MCD & $2.570\cdot10^{-2}$ & $5.891\cdot10^{-2}$ & 0.439 & $1.472\cdot10^{-6}$ & 1.230 & $2.261\cdot10^{10}$ \\ 
        SVI & $3.864\cdot10^{-2}$ & $8.234\cdot10^{-2}$ & 0.127 & $2.724\cdot10^{-2}$ & 3.339 & -$1.983\cdot10^{6}$ \\ 
        EGP & $1.319\cdot10^{-2}$ & $2.762\cdot10^{-2}$ & 0.289 & $5.005\cdot10^{-3}$ & 1.494 & -$4.435\cdot10^{6}$ \\ 
        AGP & $2.809\cdot10^{-2}$ & $5.901\cdot10^{-2}$ & 0.139 & $1.966\cdot10^{-3}$ & 2.876 & $8.840\cdot10^{6}$ \\ \hline \hline 
    \end{tabular}
    \label{tab:dlt0_metrics}
\end{table}

Table \ref{tab:dlt0_metrics} shows the various performance and uncertainty metrics for each method in the $D \le 0$ case.
Unlike the $D \ge 0$ case, the Exact GP is best in performance followed by the DE.
This is an interesting result because the $D \le 0$ is more difficult for the Approximate GP to model despite having larger space complexity.
One can also see the DE and SVI Networks out-perform the GP methods in the Miscalibration area metric.
The Negative Log-Likelihood also shows the Approximate GP method is susceptible to bad uncertainty quantification.
Where as the Deep Ensemble performs equally well on both out-of-training cases and out-performs the more computationally expensive Exact GP.
Additionally, the relatively equal accuracy and different NLL of DE and SVI is similar to the result in Izmailov \cite{izmailov2021bayesian}.
Because of this, the Deep Ensemble is overall the best method for epistemic uncertainty quantification and its results can only be improved by more sophisticated initialization techniques.

To end the discussion of the different uncertainty quantification methods for ML models, this paper gives an additional, more qualitative plot of the different methods.
To do this, one must think of an ML model in its most fundamental form, an operator that maps real-valued inputs to real-valued outputs.
Although the mapping of a ML model is defined on an infinite input space, one can characterize the function space of the posterior samples on a finite number of testing inputs.
Each posterior sample can then be visualized using dimensionality reduction methods such as PHATE (Potential of Heat-diffusion for Affinity-based Trajectory Embedding) \cite{moon2019phate}.
A method such as PHATE balances the influence of global and local structure in the reduced data to produce visualizations with more meaningful distance and directional relationships.

\begin{figure}[h]
    \centering
    \includegraphics[scale=0.75]{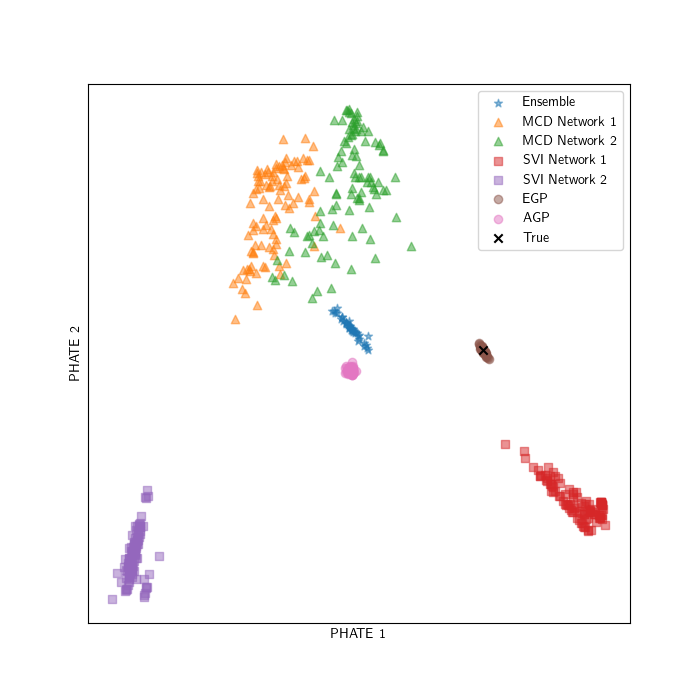}
    \caption{A plot of the PHATE dimensionality reduction embedding of each posterior sample for 2,000 test input predictions for Deep Ensembles, 2 MC-Dropout (MCD) Networks, 2 SVI Networks, Exact Gaussian Process (EGP), and Approximate Gaussian Process (AGP).}
    \label{fig:PHATE}
\end{figure}

Fig. \ref{fig:PHATE} shows the posterior approximation samples of an Ensemble, three different MC-Dropout Networks, and three different SVI Networks.
Analyzing the global structure of the plot, the accuracy hierarchy of different methods is shown by the relative distance from any point to the reference point.
This shows qualitatively the Exact GP has the highest accuracy by a considerable margin and the other methods are much closer to one another.

Analyzing the local structure in Fig. \ref{fig:PHATE}, one can relate the Epistemic uncertainty or spread of the Turbulence closure surrogates to prediction confidence.
One can see the small spread or relatively low Epistemic uncertainty in the predictions of the different GP samples.
This indicates a surrogate using an Exact or Approximate GP for uncertainty quantification will be very confident, sometimes overconfident, in its predictions of flow parameters ($G_1,G_2,G_3$).
Of the Neural Network-based Closures, the Deep Ensembles exhibit the next smallest Epistemic uncertainty.
This indicates the functions represented by each of the DE members represents a functions with similar predictions.
One can also see in Fig. \ref{fig:PHATE} that the Epistemic uncertainty of the SVI networks is lower than the MC-Dropout Networks.
However, the MC-Dropout Networks are overall closer to the reference, representing their increased accuracy over the SVI Networks.

Analyzing the global structure of Fig. \ref{fig:PHATE} shows that every Inference Method yields different functions that explain the data.
This is evident because almost none, with the exception of overlap in the MC-Dropout Networks, overlap at all with each other in the visualization.
Even in the MC-Dropout networks, the average predictions of each Network yield some separation.
Fig. \ref{fig:PHATE} illustrates pitfalls of Epistemic uncertainty quantification in NN-based surrogates using approximate methods such as the ones in this paper.
This laments the importance of accurate Epistemic uncertainty quantification because, with accurate estimates, a modeler can revert to a principled model, such as $k$-$\omega$, in the presence of high surrogate uncertainty.

\section{Conclusions}
\label{sec::conc}

This paper presents three Neural Network-based approaches and two implementations the Gaussian Process method for turbulence closure surrogate model development.
All these methods have the ability to quantify the epistemic uncertainty of the surrogate turbulence model.
The methods' performance and UQ accuracy are then evaluated for \textit{in-training} and \textit{out-of-training} cases.

In the in-training results, the Exact Gaussian process is the most accurate, followed by the Deep Ensemble across all metrics analyzed.
However, whole the performance of the Exact Gaussian Process comes at the high computational cost of inverting the covariance matrix at every step during training, the Deep Ensemble has significantly less training time due to the computational efficiencies of backpropagation.
The in-training results also show, as expected, that the Monte-Carlo Dropout, SVI, and Approximate Gaussian process methods reduce the accuracy of the prediction of the surrogate model.
For Monte-Carlo Dropout and SVI this is because their training process introduces extra noise in the gradient evaluations for SGD.
For the Approximate Gaussian process, the degradation in performance is due to the number of inducing points.

In the out-of-training cases, as expected, all methods do not generalize to the unseen training region.
Similar to the in-training case, both out-of-training cases show the Exact GP has the best accuracy, followed by Deep Ensembles. 
However, in uncertainty quantification, the Deep Ensemble exhibits uncertainty in the out-of-training region without coordination of the different members, like the one that is observed Exact Gaussian process.
This shows the Epistemic uncertainty of the Deep Ensemble is due to differing predictions of its members as opposed to the predefined uncertainty as a function of distance in the Exact GP.
In the Negative-Log-Likelihood metric, the Deep Ensemble outperforms the Exact Gaussian Process, which shows that the Gaussian Process gives inaccurate uncertainty estimates more often than the Deep Ensembles.
This leads to the conclusion that Deep Ensembles are the overall best uncertainty quantification method.

Future work will generalize these conclusions to turbulence models learned from high-fidelity Direct Numerical Simulations. Additionally, the learned surrogate models equipped with uncertainty quantification will be implemented in RANS CFD codes to propagate the model uncertainty into observables with engineering interest, such as the heat exchange coefficients or the skin friction factors.

%%%%%%%%%%%%%%%%%%%%%%%%%%%%%%%%%%%%%%%%%%%%%%%%%%%%%%%%%%%%%%%%%%%%%%%%%

%\begin{comment}
%\subsection{\label{sec:level2}Second-level heading: Formatting}

\section*{Acknowledgment}

Funding for this research has been provided by the Office of Nuclear Energy's Nuclear Engineering University Program FY-2023. Authors would like to thank the computational time on Polaris (ALCF) and Frontier (OLCF) through DOE's ALCC program, and time computational time on Utah CHPC. 

\section*{Data Availability Statement}

The data used for generating the results is available in a GitHub repository \url{https://github.com/Crashprn/UncertaintyQuantification}. 

\appendix

\section{Uncertainty and Performance Metrics}
\label{append:UQ_Perf}
\subsection{Performance Metrics}

\subsubsection{Median Absolute Error}
Median Absolute Error is a performance metric for measuring model error in a way that is insensitive to outliers.
This is due to the median operation, which largely ignores a model's lowest and highest absolute error.

\begin{equation}
    MDAE(D, f_\theta) = median(|y_i - f_\theta(x_i)|)
\end{equation}

Where $D$ is a dataset of $(x_i,y_i)$ pairs, $f_\theta$ is the model, and the median operation is evaluated over the entire dataset.

\subsubsection{Mean Absolute Error (MAE)}

Mean absolute error is a performance metric less susceptible to outliers than mean squared error.
This is because the errors are only made positive through an absolute value operation, which does not magnify errors larger than 1 like mean squared error.

\begin{equation}
    MAE(D, f_\theta) = \frac{\sum_{i=1}^N |y_i - f_\theta(x_i)|}{N}
\end{equation}

Where $D$ is a dataset of $(x_i,y_i)$ pairs, $f_\theta$ is the model, and $N$ is the size of the dataset $|D|$.

\subsubsection{Root Mean Squared Error (RMSE)}

Root mean squared error is a variant of mean squared error where the square root of the result is taken so the error has the same units as the predicted value.

\begin{equation}
    RMSE(D, f_\theta) = \sqrt{\frac{\sum_{i=1}^N (y_i - f_\theta(x_i))^2}{N}}
\end{equation}

Where $D$ is a dataset of $(x_i,y_i)$ pairs, $f_\theta$ is the model, and $N$ is the size of the dataset $|D|$.

\subsubsection{Mean Absolute Relative Percent Difference (MAEPD)}
Mean absolute relative percent difference\cite{tran2020methodscomparinguncertaintyquantifications} is a performance metric measuring the percent difference of the error of a prediction compared to the predicted quantity.
One may notice it compares the distance of the error with respect to the distances of the true and predicted value, where distance is measured using an absolute value.

\begin{equation}
    MARPD(D,f_\theta) = \frac{100}{N} \sum_{i=1}^N \frac{|y_i - f_\theta(x_i)|}{|y_i| +  |f_\theta(x_i)|}
\end{equation}

Where $D$ is a dataset of $(x_i,y_i)$ pairs, $f_\theta$ is the model, and $N$ is the size of the dataset $|D|$.

\subsubsection{$R^2$ Score}

The $R^2$ score or Coefficient of Determination is a performance that compares the variance of a true value with a prediction to the variance of the true value with its mean.
This effectively measures how much variance is eliminated by the model.

\begin{equation}
    R^2(D, f_\theta) = 1 - \frac{\sum_{i=1}^N (y_i - f_\theta(x_i))^2}{\sum_{i=1}^N (y_i - \Bar{y})^2}
\end{equation}

Where $D$ is a dataset of $(x_i,y_i)$ pairs, $f_\theta$ is the model, $\Bar{y}$ is the mean output value, and $N$ is the size of the dataset $|D|$.

\subsection{Uncertainty Metrics}

\subsubsection{Empirical vs Expected Cumulative Density Function (CDF)}

In a regression problem, it is important to compare the distribution over the outputs of a model in terms of a well-defined distribution, such as a standard Gaussian.
This can be done by leveraging the Multivariate Central Limit Theorem (MCLT), used in statistical hypothesis tests.
Let $Y_x$ be a random vector conditioned on the value of x and let $f_\theta(x)$ model the mean value of the distribution of $Y_x$.
Then, by the MCLT subtracting the mean value from $Y_x$ over multiple samples should converge to a zero mean MV-Gaussian with some covariance matrix.
Additionally, when the elements of $Y_x$ are assumed to be uncorrelated with one another, one can simply divide each component by the standard deviation to yield standard normal random variables.
One can then compare how well a model $f_\theta(x)$ estimates the mean of $Y_x$ by comparing the empirical Cumulative Density Function (CDF) with a standard normal.
Or, inversely, one can use the inverse CDF of a standard normal to generate theoretical probability bounds for the empirical distribution of $Y_x$.

\begin{equation}
    F^{-1}_{Y_x}(p) = diag(\sigma^2(x))F^{-1}_Z(p) + f_\theta(x) 
\end{equation}

The inverse CDF of $Y$ can then be used to determine the empirical CDF of Y using $m$ confidence intervals $0 < p_1 < ... < p_m \le 1$.

\begin{equation}
    \hat{p}_j = \frac{\left|\{y_t: y_t \le F^{-1}_{Y|X}(p_j|x_t), t = 1, ..., T\}\right|}{T}
\end{equation}

\subsubsection{Miscalibration Area Error (MisCal)}

The miscalibration area error between an expected and empirical CDF measures the total area between a diagonal line $(p_j,p_j)$ and the calibration curve $(p_j, \hat{p}_j)$.
This can be accomplished using any integration scheme but this paper uses trapezoidal integration.

\begin{equation}
    MisCal(p,\hat{p}) = \frac{1}{2}\sum_{i=1}^m (|\hat{p}_i - p_i| + |\hat{p}_{i+1} - p_{i+1}|) \delta p
\end{equation}

Where $p$ is the set of m confidence intervals, $\hat{p}$ is the set of m empirical confidence intervals, and $\delta p$ is a chosen step size for the m confidence intervals.

\subsubsection{Calibration Error (CE)}

The calibration error, found in Kuleshov \cite{Kuleshov2018a}, is the squared difference between the expected and empirical confidence intervals.

\begin{equation}
    CE(p,\hat{p}) = \sum_{i=1}^m \omega_i (p_i - \hat{p}_i)^2
\end{equation}

Where $p$ is the set of m confidence intervals, $\hat{p}$ is the set of m empirical confidence intervals, and $\omega_i$ is some weight on the squared errors.
Kuleshov \cite{Kuleshov2018a} suggests picking $\omega_i$ corresponding to the fraction of points in that confidence interval.

\subsubsection{Sharpness Score (Sha)}

The sharpness score is a measure of the average standard deviation of all outputs.

\begin{equation}
    Sha(D, f_\theta) = \sqrt{\frac{1}{T} \sum_{i=1}^T \sigma^2(x_i)}
\end{equation}

Where T is the size of the dataset and $\sigma^2(x_i)$ is the sample variance of $f_\theta(x_i)$.
For vector outputs, one can calculate a scalar value by using the magnitude of the sharpness.

\subsubsection{Coefficient of Variation ($C_v$)}

Tran\cite{tran2020methodscomparinguncertaintyquantifications} defines the coefficient of variation as a measure of the spread of the variance compared to the average variance.

\begin{equation}
    C_v(D, f_\theta) = \frac{\sqrt{\frac{\sum_{i=1}^N (\sigma(x_i) - \mu_\sigma)^2}{N-1}}}{\mu_\sigma}
\end{equation}

Where $\sigma(x_i)$ is the standard deviation of the prediction $f_\theta(x_i)$ and $\mu_\sigma$ is the average standard deviation over the entire dataset.

\subsubsection{Negative Log-Likelihood (NLL)}

The negative log-likelihood of a dataset is a strictly proper scoring rule that accounts for the both the accuracy and uncertainty quality of a model.
The NLL assumes a Independent and Identically Distributed (IID) dataset $D$ and is defined by an assumed likelihood function.
This paper assumes a Multivariate Gaussian likelihood with a $f_\theta$ as the mean and a diagonal covariance $\Sigma_i = diag(\sigma^2(x_i))$.

\begin{equation}
    NLL(D, f_\theta) = \frac{1}{2}\sum_{i=1}^N\left( d\log(2\pi) + \log(|\Sigma_i|) + (y_i - f_\theta(x_i))^T\Sigma_i^{-1}(y_i - f_\theta(x_i))\right)
\end{equation}

Where $d$ is the dimensionality of the output, $|\Sigma|$ is the determinant of the covariance, and N is the number of data points.

\section{Supplementary Tables}

\begin{table}[H]
    \centering
    \setlength{\tabcolsep}{5pt}
    \renewcommand{\arraystretch}{1.3}
    \caption{A table of the performance and uncertainty metrics on the full input space for Deep Ensembles (DE), Monte-Carlo Dropout (MCD), Stochastic Variational Inference (SVI), Exact Gaussian Process (EGP), and Approximate Gaussian Process (AGP).}
    \begin{tabular}{c|cccccc} 
        Method & MAE & RMSE & MisCal & Sha & $C_v$ & NLL \\ \hline \hline
        DE & $1.933\cdot10^{-4}$ & $4.590\cdot10^{-4}$ & 0.096 & $2.830\cdot10^{-7}$ & 1.463 & -$1.042\cdot10^{7}$ \\ 
        MCD & $4.016\cdot10^{-4}$ & $8.450\cdot10^{-4}$ & 0.143 & $4.726\cdot10^{-6}$ & 2.546 & -$9.248\cdot10^{6}$ \\ 
        SVI & $5.549\cdot10^{-4}$ & $8.758\cdot10^{-4}$ & 0.131 & $1.725\cdot10^{-6}$ & 0.139 & -$8.419\cdot10^{6}$ \\ 
        EGP & $9.734\cdot10^{-7}$ & $2.145\cdot10^{-5}$ & 0.397 & $1.141\cdot10^{-7}$ & 0.341 & -$1.075\cdot10^{7}$ \\ 
        AGP & $9.805\cdot10^{-4}$ & $1.679\cdot10^{-3}$ & 0.341 & $8.015\cdot10^{-5}$ & 0.388 & -$6.119\cdot10^{6}$ \\ \hline \hline
    \end{tabular}
    \label{tab:full_metrics}
\end{table}

\bibliography{aipsamp}% Produces the bibliography via BibTeX.

\end{document}